\documentclass{article}

\usepackage{PRIMEarxiv}

\usepackage[utf8]{inputenc} 
\usepackage[T1]{fontenc}    
\usepackage{hyperref}       
\usepackage{url}            
\usepackage{booktabs}       
\usepackage{amsfonts}       
\usepackage{nicefrac}       
\usepackage{microtype}      
\usepackage{lipsum}
\usepackage{graphicx}       
\graphicspath{{media/}}     
\usepackage{natbib}
\usepackage[linesnumbered,ruled,vlined]{algorithm2e} 

\usepackage{microtype}
\usepackage{graphicx}
\usepackage{subfigure}
\usepackage{booktabs} 
\usepackage{tikz}
 \usetikzlibrary {arrows.meta}
\usepackage{multirow}
\usepackage{hhline}

\usepackage{pifont}

\usepackage{siunitx}
\usepackage{hyperref}

\usepackage{amsmath}
\usepackage{amssymb}
\usepackage{mathtools}
\usepackage{amsthm}

\theoremstyle{plain}
\newtheorem{theorem}{Theorem}[section]
\newtheorem{proposition}[theorem]{Proposition}
\newtheorem{lemma}[theorem]{Lemma}
\newtheorem{corollary}[theorem]{Corollary}
\theoremstyle{definition}

\theoremstyle{remark}

\newcommand{\argmax}{\operatorname{arg\max}}

\newcommand{\indep}{\perp \!\!\! \perp}

\title{Learning Causal Response Representations through Direct Effect Analysis}

\author{
  Homer Durand, Gherardo Varando, Gustau Camps-Valls \\
  Image and Processing Lab \\
  Universitat de Valencia \\
  Valencia\\
  \texttt{homer.durand@uv.es} \\
  \texttt{gherardo.varando@uv.es} \\
  \texttt{gustau.camps@uv.es}
}

\begin{document}
\maketitle

\begin{abstract}
 We propose a novel approach for learning causal response representations. Our method aims to extract directions in which a multidimensional outcome is most directly caused by a treatment variable. By bridging conditional independence testing with causal representation learning, we formulate an optimisation problem that maximises the evidence against conditional independence between the treatment and outcome, given a conditioning set. This formulation employs flexible regression models tailored to specific applications, creating a versatile framework. The problem is addressed through a generalised eigenvalue decomposition. We show that, under mild assumptions, the distribution of the largest eigenvalue can be bounded by a known $F$-distribution, enabling testable conditional independence. We also provide theoretical guarantees for the optimality of the learned representation in terms of signal-to-noise ratio and Fisher information maximisation. Finally, we demonstrate the empirical effectiveness of our approach in simulation and real-world experiments. Our results underscore the utility of this framework in uncovering direct causal effects within complex, multivariate settings.
 
\end{abstract}

\section{Introduction}\label{sec:intro}

    Representation learning has been a foundational tool in modern machine learning, enabling models to automatically extract features from high-dimensional data~\citep{bengio2013representation,lecun2015deep}. However, traditional approaches often fail to capture the causal mechanisms that underlie data generation, leading to poor generalisation under data distribution shifts. To address these shortcomings, \emph{causal representation learning} (CRL) has emerged as a crucial approach to integrate causality into representation learning~\citep{scholkopf2021toward}. By learning representations that reflect the causal structure of the data, models can become more robust to distribution shifts and provide better causal insights for downstream tasks. This enables the modelling of intervention effects and the construction of counterfactuals, allowing for the analysis of questions that classical statistical models may struggle with, such as estimating the effects of policies. 
  
    A key focus of causal inference literature is understanding how variables influence one another along different pathways~\citep{pearl2014interpretation}. Of particular interest to this work is the direct effect of a cause on an outcome variable while controlling for confounders and mediators~\citep{pearl2022direct}. Mediators transmit the effect of the cause to the outcome, while confounders influence both the cause and the outcome. Studying direct effects rather than total effects is essential for several reasons. For instance, it allows isolating specific mechanisms in science, such as assessing the effect of greenhouse gas emissions on local temperature while controlling for natural climate variations (that emissions may also influence). Finally, it helps disentangle immediate effects from delayed downstream effects, which may have a longer-term impact on the outcome. 

    When the outcome is multi-dimensional, identifying a subspace where the causes maximally influence it can benefit various tasks. In these cases, it is often impractical to observe how each dimension's distribution is shifted by the intervention. Therefore, it is of interest to examine this shift in a lower-dimensional space (e.g., 1-D or 2-D). This approach can also help discover simple, low-dimensional representations that capture relevant information about the intervention's effect. Additionally, it can help disentangle the direction in which the outcome is affected by the intervention from the direction where the distribution remains unchanged. We will demonstrate that this has important implications in different application domains with a focus on climate change attribution. 

    While considerable work has focused on learning representations for confounder adjustment in causal effect estimation~\citep{louizos2017causal}, modelling representations of causes~\citep{Arjovsky2019, Peters2016}, or uncovering latent causal graphs~\citep{locatello2019challenging}, the representation of effects remains largely underexplored. Here, we aim to bridge this gap by learning a mapping of the response variable through the maximisation of a conditional independence statistic. Under certain structural assumptions, the method identifies the direction in which the effect of interventions is most observable. By using conditional expectation estimators, it adapts to different data types through various regression models.
    


\section{Preliminary}

            

Let $X \in \mathbb{R}^p$, $Y \in \mathbb{R}^d$, and $Z \in \mathbb{R}^r$ be three random vectors  with density function $p(x, y, z)$ and assume that their joint distribution is absolutely continuous with respect to the Lebesgue measure. We also assume that $X$ and $Z$ are known causes of $Y$, but the relation between $X$ and $Z$ is left unspecified, allowing it to be a confounder, a mediator, or both. 
We aim to identify the component of $Y$ that is most directly caused by $X$, by finding $\mathbf{w}$ that maximises the causal relationship between $X$ and $\mathbf{w}^\top Y$. In the following, we clarify key terms related to the concepts of direct effect and conditional independence.
We begin by considering James Woodward's \textit{manipulationist} definition~\citep{woodward2005making} of a direct cause:
\begin{quote}
    \textit{A necessary and sufficient condition for $X$ to be a direct cause of $Y$ with respect to some variable set [$Z$] is that there be a possible intervention on $X$ that will change $Y$ (or the probability distribution of $Y$) when all other variables in [$Z$] besides $X$ and $Y$ are held fixed at some value by interventions.}
\end{quote}
The distribution of $Y$ under intervention is called the direct effect (DE) of $X$ on $Y$. For simplicity, we assume that the effects of $X$ and $Z$ on $Y$ are additive, as formalised in the model assumption in Sec. \ref{sec:Theory}. Thus, DE can be written as:
\begin{align}\label{eq:DE}
    DE(x) &= p(Y|do(X=x), do(Z=z)).
\end{align}
The variable $Y$ under the intervention $do(X = x)$ is denoted as $Y^x$. 
We note that under the assumption of additivity, the direct effect is equivalent to the natural direct effect~\citep[see][section 4.5]{pearl2009causality}. In some contexts, it is described in terms of conditional expectation—referred to as the expected direct effect (EDE)~\citep[see][section 4.5.4]{pearl2009causality}. However, we avoid this reduction, as the (conditional) expectation masks valuable information needed to identify the direction in which $Y$ is most caused by $X$, namely $Y$'s noise structure. Additionally, the term \textit{Gradient DE} (GDE) will be used to denote the vector of the partial derivative of Eq. \eqref{eq:DE} with respect to $x$, capturing how small variations in the intervention affect $Y$.  
In some cases, the Gradient Direct Effect (GDE) lies in a subspace $\mathbb{R}^q \subset \mathbb{R}^d$, meaning that the distribution $P(Y|do(X=x))$ is affected by the intervention only in this subspace, while the remaining dimensions of the space are unaffected. We refer to this as the direct effect subspace (DES). Our work focuses on recovering this reduced space, with its basis ordered by the variance in $Y$ explained by $X$, while controlling for $Z$, analogous to how Principal Component Analysis (PCA) identifies directions of maximum variance in a random vector.

We now summarise conditional independence testing, which plays an important role in our work. 
%
%
%
We say that $X$ is conditionally independent of $Y$ given $Z$, denoted $X \indep Y \mid Z$, if for all $x \in \mathbb{R}^p$, $y \in \mathbb{R}^d$, and $z \in \mathbb{R}^r$, $p(y \mid x, z) = p(y \mid z)$, or equivalently, $p(x, y \mid z) = p(x \mid z)p(y \mid z)$. This means that, given $Z$, $X$ adds no additional information about $Y$. Let $P$ denote the joint distribution of $(X, Y, Z)$ such that $P \in \mathcal{P}$ if $X \indep Y \mid Z$ holds (the null hypothesis), and $P \in \mathcal{Q}$ if $X \not \indep Y \mid Z$ (the alternative hypothesis). A Conditional Independence Test (CIT) is formulated as $H_0:   P \in \mathcal{P} \quad \text{vs.} \quad  H_1:   P \in \mathcal{Q}$.
Given i.i.d. observations $\mathbf{X} \in \mathbb{R}^{n \times p}$, $\mathbf{Y} \in \mathbb{R}^{n \times d}$, and $\mathbf{Z} \in \mathbb{R}^{n \times r}$, we use a statistic $\mathbf{T}_n(\mathbf{X}, \mathbf{Y}, \mathbf{Z})$, and reject $H_0$ when $\mathbf{T}_n$ deviates sufficiently from its expected distribution under $P \in \mathcal{P}$. 


\subsection{Introductory example}\label{sec:example}

\begin{figure*}
    \centering
    \includegraphics[width=1\linewidth]{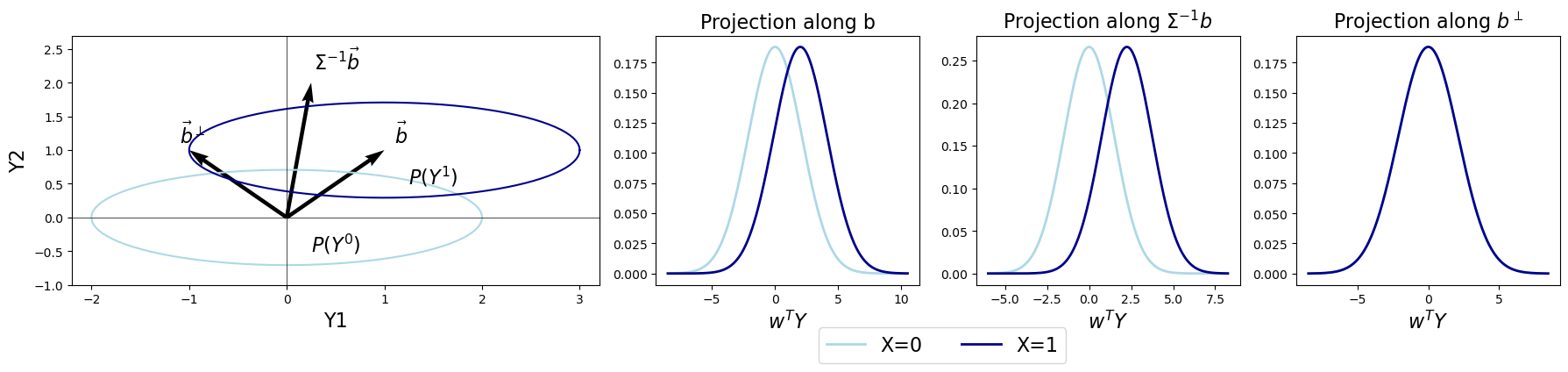}
    \caption{Illustration of the linear model from Sec.~\ref{sec:example} with $\mathbf{b} = (1,1)^\top$ and $\boldsymbol{\Sigma} = ( 4, 0 ; 0, 1/2 )$, showing the one-sigma ellipsoid for $Y^0$ and $Y^1$. For one-dimensional $X$, $Y^x$ shifts along $\mathbf{b}$, but projection along $\mathbf{b}$ is suboptimal. In contrast, projection along $\boldsymbol{\Sigma}^{-1} \mathbf{b}$ is optimal, with $(\boldsymbol{\Sigma}^{-1} \mathbf{b}, \mathbf{b}^\perp)$ forming a natural basis for the intervention space, where the first axis captures the intervention effect and the second contains no information.
}
    \label{fig:optimal_proj}
\end{figure*}


Through a simple example, we demonstrate that EDE is generally suboptimal for distinguishing $Y$ distributions under different interventions and that strategically maximising a CIT statistic may be more effective.
Let us consider the simple linear model $Y = \mathbf{b} X + \mathbf{c} Z + N$
with $Y \in \mathbb{R}^d$, $X \in \mathbb{R}$, $Z \in \mathbb{R}^p$, and $N \in \mathbb{R}^d$. Let also $\boldsymbol{\Sigma}$ denote the covariance matrix of $Y^x$. The relationship between $X$ and $Z$ is not relevant in this context, as we focus on the intervention distribution $Y^x$, and such an intervention breaks the statistical association between $X$ and $Y$.

In a linear model, the Gradient EDE is given by the weight vector $\mathbf{b}$ for interventions on $X$~\citep[][ex. 6.42]{peters2017elements}, often called the direct effect. This means that the distribution of $Y$ is only shifting along the $\vec{\mathbf{b}}$ axis when intervening on $X$ (see Fig. \ref{fig:optimal_proj}$)$. The most common approach to find $\mathbf{b}$ is by analysing the weights of $X$ in the conditional expectation $\mathbb{E}[Y | X, Z]$. Alternatively, $\mathbf{b}$ can be obtained by maximising the partial correlation between $X$ and $\mathbf{w}^\top Y$ given $Z$. When $N$ is isotropic, the vector $\mathbf{w}$ that maximises this partial correlation is indeed $\mathbf{b}$. Since partial correlation is used in CITs, through Fisher's $Z$ transformation~\citep{fisher1915frequency}, which applies the \textit{arctanh} function to the partial correlation, the Gradient EDE can be recovered by finding the direction that maximises a CIT statistic for $\mathbf{w}^\top Y$.


However, when the noise $N$ is non-isotropic, $\mathbf{b}$ may not be optimal for isolating the causal effect of $X$ on $Y$. In this case, the direction $\mathbf{b}$ may align too closely with the noise structure of $Y$, making the intervention's effect less discernible. While regression-based approaches fail to account for the noise structure, CIT statistics balance signal detection (the effect of the intervention) and noise reduction to obtain optimal power. For non-isotropic noise, it can be shown that the most discriminative direction for the intervention is $\boldsymbol{\Sigma}^{-1} \mathbf{b}$. While EDE generally fails, maximising the partial correlation recovers this optimal direction. This illustrates how identifying the direction in which $Y$ maximises a conditional independence statistic can effectively uncover the subspace of $Y$ most caused by $X$.

This example is illustrated in Fig. \ref{fig:optimal_proj}, where we observe that projection along $\boldsymbol{\Sigma}^{-1} \mathbf{b}$ improves the separability of distributions under different interventions (here $X=0$ and $X=1$). A natural basis for representing interventions on $Y$ is then $(\boldsymbol{\Sigma}^{-1} \mathbf{b}, \mathbf{b}^\perp)$, where the first vector captures all information about the intervention, and the second contains no information. These axes need not be orthogonal. Under favorable conditions, such as a rapid decay in the eigenvalues of the covariance matrix $\boldsymbol{\Sigma}$, the noise in the distribution of $Y$ along the optimal direction $\boldsymbol{\Sigma}^{-1} \mathbf{b}$ diminishes as the dimensionality increases, concentrating the distribution’s mass in a single point and achieving optimal separability of the intervention distributions.

\subsection{Related work}
Although the idea of learning representations of effects of causes is, to our knowledge, novel, there are important connections between our work and other fields. 
It intersects two key areas of statistical learning: conditional independence testing and causal representation learning. Below, we summarise the most relevant results in these and other relevant fields.

\textbf{Conditional Independence Testing:} A variety of methods address this problem, broadly classified into nonparametric and parametric approaches. Nonparametric methods, like kernel-based tests~\citep{zhang2011kernel}, nearest-neighbour methods~\citep{runge2018conditional}, and mutual information-based tests~\citep{fukumizu2008kernel}, offer flexibility but are computationally expensive. Regression-based approaches~\citep{Shah2018TheHO} test residual dependencies or whether $X$ improves prediction of $Y$ given $Z$~\citep{chow1960tests}. Parametric methods, such as partial canonical correlation analysis (CCA)~\citep{Rao1969}, assume linearity and Gaussianity, providing computational efficiency at the cost of strong assumptions. While these methods balance complexity, power, and robustness, they do not explicitly recover an optimal subspace for testing, though they may indirectly solve an optimisation problem that achieves this, as we will demonstrate.

\textbf{Causal Representation Learning (CRL):}  CRL~\citep{scholkopf2021toward} aims to learn representations that capture causal mechanisms, enhancing generalisation, interpretability, and robustness. Leveraging invariance across environments~\citep{Arjovsky2019}, recent methods focus on learning representations for confounders or predictors to estimate causal effects~\citep{yao2018representation, Yang2021, locatello2019challenging}, with some extending to temporal data~\citep{lachapelle2022disentanglement, lippe2022causal}. While prior work targets confounder or predictor representations, our method focuses on causal effect representation of the outcome, filling a gap in previous approaches.

\textbf{Connections to Signal Detection:} Our framework relates to signal detection~\citep{macmillan2002signal, Kay1998Detection, kay1993fundamentals}, aiming to identify a deterministic signal \( X \) in noisy observations \( Y = X + N \). In climate science, this is addressed by the ``optimal fingerprint''~\citep{hasselmann1993optimal}, which maximises the signal-to-noise ratio of a linear projection of observations. This enables a direct test for the detection of climate change while recovering a useful climate pattern.

\textbf{Sufficient Dimensionality Reduction (SDR):} There are also similarities with the SDR framework~\citep{globerson2003sufficient, fukumizu2009kernel}, which aims to find a sufficient statistic $\mathbf{w}^\top X$ such that $p(Y|X) = p(Y|\mathbf{w}^\top X)$. The reduced space thus containing all relevant information in $X$ to predict $Y$. Our work focuses on finding a sufficient statistic specifically for the DE, to know, a subspace that retains all relevant information about the DE.

\section{Learning Framework}\label{sec:LF}

Our goal is to identify the components of $Y$ that are most caused by $X$, conditional on $Z$, assuming all confounders $C \subseteq Z$ are observed and the causal relationship $X \to Y$ is known. Specifically, we aim to find a subspace of $Y$ that encapsulates all information about interventions on $X$. To achieve this, we represent the subspace as a linear transformation, $\tilde{Y} = \mathbf{W}^\top Y \in \mathbb{R}^q$, where $\mathbf{W} \in \mathbb{R}^{d \times q}$. For simplicity, we focus on the case where $q = 1$, and identify a vector $\mathbf{w} \in \mathbb{R}^d$ such that $\mathbf{w}^\top Y \in \mathbb{R}$ captures the maximum amount of information that a one-dimensional representation of $Y$ can convey about the intervention on $X$. The case for $q > 1$ is discussed in Section \ref{sec:more_components}.

\subsection{Maximisation of a CIT statistic}

We propose a class of learning algorithms that maximise a CIT statistic to find $\mathbf{w}$, 
following the optimisation problem:
\begin{align}\label{eq:optimisation_problem}
    \mathbf{w}^\star = \argmax_{\mathbf{w}} T(X, \mathbf{w}^\top Y, Z).
\end{align}
Here, $X$, $\mathbf{w}^\top Y$, and $Z$ are treated as random variables, as we consider a \textit{population} version of the test statistic. This formulation provides theoretical guarantees for recovering the latent structure (see Sec. \ref{sec:latent_structure_recovery}) and the optimality of the learned representation in terms of Fisher information. We denote $T$ the population loss and $\mathbf{T}_n$ its empirical counterpart.

Building on this idea, we propose a flexible framework based on nested predictive models of $Y$.
This approach assesses conditional independence by analysing the residuals from two regression models. The restricted model regresses $Y$ on $Z$ alone, while the full model includes both $X$ and $Z$. Conditional independence is evaluated by comparing the residuals of these models, without assuming a specific functional form between $X$ and $Y$. This flexibility makes the framework broadly applicable across various settings, accommodating complex, nonlinear relationships between variables.
Let us define $ R_{\text{full}}^2(\mathbf{w}) = \mathbb{E}\left[(\mathbf{w}^\top Y - \mathbb{E}[\mathbf{w}^\top Y | X, Z])^2\right] $ and $ R_{\text{res}}^2(\mathbf{w}) = \mathbb{E}\left[(\mathbf{w}^\top Y - \mathbb{E}[\mathbf{w}^\top Y | Z])^2\right] $ as the population mean squared error when predicting $ \mathbf{w}^\top Y $ from the full model (including both $X$ and $Z$) and the restricted model (including only $Z$), respectively. A straightforward way of enforcing conditional dependence--maximising the \textit{distance} between $p(y | x, z)$ and $p(y | z)$--is to maximise the distance between the residuals of the full regression model and the restricted one. This leads to the simple loss function:
\begin{align}\label{eq:simple_population_loss}
    T_S(X, Y, Z; \mathbf{w}) = R_{\text{res}}^2(\mathbf{w}) - R_{\text{full}}^2(\mathbf{w}).
\end{align}
Under the null hypothesis, both regression models have equal predictive power, but the full model, with more degrees of freedom, yields smaller residuals. This can also be viewed through an information theory perspective, detailed further in Sec. \ref{sec:IT_and_SNR}.
%
%
However, this loss function is unbounded with respect to $\mathbf{w}$; thus, it is necessary to impose additive constraints on $\mathbf{w}$ to avoid trivial solutions. The most straightforward way to constrain the loss is to limit $\mathbf{w}$ to be a unit norm vector, i.e., $\|\mathbf{w}\| = 1$. We show in Lemma \ref{lemma:EGV_sol} in supplementary material that this approach recovers the EDE and is thus suboptimal for non-isotropic noises.

Another approach, is to constrain the full residuals to be fixed, leading to the following loss function:
\begin{align}\label{eq:population_loss_F}
    T_{F}(X, Y, Z; \mathbf{w}) = \frac{ R_{\text{res}}^2(\mathbf{w}) - R_{\text{full}}^2(\mathbf{w})}{R_{\text{full}}^2(\mathbf{w})}.
\end{align}
In the context of a linear Gaussian SCM, this statistic can be interpreted as an F-test between nested models (aka Chow test~\citep{chow1960tests}), which is commonly used for variable selection~\citep{hocking1976biometrics} or causal discovery~\citep{nogueira2022methods}. When the conditioning set $Z$ consists of the past values of $Y$, the empirical version of $T_F$ corresponds to the statistic of the well-known Granger causality test~\citep{granger1969investigating}. In this context, the maximisation of $T_F$ with respect to $\mathbf{w}$ leads to a causal representation method known as Granger PCA~\citep{varando2022learning}. This further emphasises how maximising a conditional independence testing statistic can be leveraged to uncover the direction in which $Y$ is most strongly caused by $X$.
Another possible constraint is grounded in detection theory~\citep{macmillan2002signal, kay1993fundamentals}. Considering that $Y$ can be decomposed into a signal term $S$ (variance related to $X$) and a noise term $N$ (variance related to $Z$ and $Y$'s intrinsic noise), we constrain the variance of $\mathbf{w}^\top N$. Assuming that the signal and noise are additive in $Y$, this constraint relates to constraining $R^2_{\text{noise}} = \mathbb{E}[(\mathbf{w}^\top Y - \mathbb{E}[\mathbf{w}^\top Y \mid X, Z=0])^2]$. 
We thus propose the  loss function:
\begin{align}\label{eq:population_loss_detect}
    T_{D} = \frac{ R_{\text{res}}(\mathbf{w})^2 - R_{\text{full}}(\mathbf{w})
     ^2}{R^2_{\text{noise}}(\mathbf{w})}.
\end{align}
It will be shown in Sec. \ref{sec:latent_structure_recovery} that this formulation is optimal under certain structural assumptions.

Canonical Correlation Analysis (CCA) \citep{hotelling1992relations} and its partial variant \cite{Rao1969} also seek a subspace that captures reduced information between $X$ and $Y$ (conditioning on $Z$ in partial CCA), enabling (conditional) independence testing. In Sec.~\ref{supp:pCCA}, we demonstrate that partial CCA aligns with our framework by interpreting it as the maximisation of a conditional independence statistic.

\subsection{Empirical estimators}\label{sec:emp_est}

We now present the practical optimisation procedure to estimate $\mathbf{w}^\star$.
Given observation (or design) matrices $ \mathbf{X} \in \mathbb{R}^{n \times p} $, $ \mathbf{Y} \in \mathbb{R}^{n \times d} $, and $ \mathbf{Z} \in \mathbb{R}^{n \times r} $, we now present empirical estimators for $\mathbf{w}_S$, $\mathbf{w}_F$ and $\mathbf{w}_D$.

Similarly, we assume that we have two estimators $ \hat{g}_{\text{full}}(X, Z) $ and $ \hat{g}_{\text{res}}(Z) $ for the conditional expectations $ \mathbb{E}[Y \mid X, Z] $ and $ \mathbb{E}[Y \mid Z] $, respectively. The learning algorithms employed to estimate these conditional expectations are not restricted, allowing users to tailor them based on their assumptions about the relationships within the data and their prior knowledge. We denote by $ \hat{\boldsymbol{\Sigma}}_{\text{full}} $, $ \hat{\boldsymbol{\Sigma}}_{\text{res}} $, and $ \hat{\boldsymbol{\Sigma}}_{\text{noise}} $ the sample covariance matrices of the residuals from the full and restricted models, as well as the noise covariance.
The three population losses can be maximised by solving the general eigenvalue problem $\hat{\mathbf{M}} \mathbf{w} = \lambda \hat{\mathbf{N}} \mathbf{w}$, where $\hat{\mathbf{M}} = \hat{\boldsymbol{\Sigma}}_{\text{res}} - \hat{\boldsymbol{\Sigma}}_{\text{full}}$ and $\mathbf{N}$ corresponds to the constraints on $\mathbf{w}$: $\hat{\mathbf{N}} = \mathbf{I}$ for $T_S$, $\hat{\mathbf{N}} = \hat{\boldsymbol{\Sigma}}_{\text{full}}$ for $T_F$, and $\hat{\mathbf{N}} = \hat{\boldsymbol{\Sigma}}_{\text{noise}}$ for $T_D$.
Given random realisations of $(X, Y, Z)$, the population matrices $\mathbf{M}$ and $\mathbf{N}$ are random, typically following a Wishart distribution. Under this condition, the first eigenvalue of the GEV problem, denoted by $\Lambda_1$, is also random. Upon observing data $(\mathbf{X}, \mathbf{Y}, \mathbf{Z})$, the empirical covariances $\hat{\mathbf{M}}$ and $\hat{\mathbf{N}}$ are fixed, and we obtain a realisation $\lambda_1 \sim \Lambda_1$ with corresponding eigenvector $\mathbf{w}_1$. We denote the eigen-pairs $(\lambda_S, \mathbf{w}_S)$, $(\lambda_F, \mathbf{w}_F)$, and $(\lambda_D, \mathbf{w}_D)$ as those corresponding to the first eigenvalues for the losses $T_S$, $T_F$, and $T_D$, respectively.

The convergence properties of these estimators are presented in Th. \ref{prop:cv} in the supplementary materials. Additional details on the estimation of the conditional expectations, as well as the estimation of other components and the stability of the solution, can be found in Section \ref{sec:Estim_details}.

\section{Theoretical guarantees}\label{sec:Theory}

In this section, we discuss the theoretical properties of the maximisation of the statistics introduced earlier. We consider the distribution of $ (X, Y, Z) \sim P $ entailed within the following Structural Causal Model (SCM):
\begin{align}\label{eq:scm}
    Y &:= \mathbf{b} \phi(X) + \psi(Z) + N_y,
\end{align}
where $ \phi(x): \mathbb{R}^p \to \mathbb{R} $, $ \psi(z): \mathbb{R}^r \to \mathbb{R}^d $, $\mathbf{b} \in \mathbb{R}^{d}$ and with $N_y \sim \mathcal{N}(0, \mathbf{\Sigma})$. Again, the relationship between $X$ and $Z$ is left undefined as applying $do(X)$ breaks any statistical dependencies that existed in the observational setting. We denote by $\mathbf{\Sigma}_{\psi(z)}$ the covariance of $\psi(Z)$. For the remainder of this section, we assume that the intervention $\phi(x)$ is bounded. 
Concretely, we assume that $X$ goes threw an information bottleneck of dimension one. The vector $\mathbf{b}$ thus gives the direction of the causal effect as intervention on $X$ will shift along axis $\mathbf{b}$. Note that if $\phi(x)$ is linear, it corresponds to the Gradient EDE.
%
All the proofs are given in Sec. \ref{sec:Proofs} in supplementary materials.

\subsection{Causal effect representation}\label{sec:latent_structure_recovery}

To better understand the properties of the different learning algorithms, it is useful to decompose the intervention distribution \( Y^x \) into a signal term and a noise term $Y^x = S(x) + N$ where \( S(x) = \mathbf{b} \phi(x) \) represents the EDE, a non-random component of \( Y^x \), and the noise term is given by \( N = \psi(Z) + N_y \), which remains random. We define the SNR of the transformed variable \( \mathbf{w}^\top Y^x = \mathbf{w}^\top S(x) + \mathbf{w}^\top N \) as  
\begin{align}\label{eq:SNR}
    \gamma^2(\mathbf{w}) = \frac{(\mathbf{w}^\top S(x))^2}{\mathbf{w}^\top \mathbf{\Sigma}_N \mathbf{w}},
\end{align}
where \( \mathbf{\Sigma}_N \) is the covariance matrix of the noise term. Notably, when the conditioning set \( Z \) is accounted for, the noise covariance \( \mathbf{\Sigma}_N \) simplifies to \( \mathbf{\Sigma} \). In this case, the optimality results that will be established for \( \mathbf{w}_D \) also apply to \( \mathbf{w}_F \).  
We now present some optimality results related to the SNR. This metric is tied to an optimal representation because, as the SNR increases, the distribution becomes more concentrated around the signal $S(x)$ Thus, the direction that maximises the SNR is the one for which small perturbations of the intervention are most observable. We thus say that a weight vector $\mathbf{w}$ is optimal if it maximises $\gamma^2(\mathbf{w})$. For general noise structures, $\mathbf{w}_D$ is shown to be optimal.

\begin{proposition}[General optimality]\label{prop:snr_max}
    Assuming $P$ is entailed in the SCM in \eqref{eq:scm}, we have that $\mathbf{w}_D$ is optimal.
\end{proposition}
%
Under stronger assumptions -- isotropy of the noises -- both $\mathbf{w}_S$ and $\mathbf{w}_F$ are shown to be optimal.
\begin{proposition}[Optimality under isotropic noise]\label{prop:snr_max_others}
    Assuming that $P$ is entailed in the SCM in \eqref{eq:scm} and that $\mathbf{\Sigma}_N$ is isotropic, we have that both $\mathbf{w}_S$ and $\mathbf{w}_D$ are optimal. Moreover, if $\mathbf{\Sigma}$ is also isotropic, then $\mathbf{w}_F$ is also optimal.
\end{proposition}
This proposition implies that when the effects of $X$ and $Z$ are assumed to be separable, $\mathbf{w}_D$ is optimal in the sense that it maximises the SNR. 
%

We now present different guarantees for the learned representation, demonstrating that in the large-dimensional regime, and under specific conditions on the characteristics of \( \mathbf{b} \), \( \boldsymbol{\Sigma} \), and \( \boldsymbol{\Sigma}_{\psi(z)} \), the signal-to-noise ratio improves as the dimensionality of \( Y \) increases, such that the signal of \( \mathbf{w}^\top Y \) completely dominates its noise.

\begin{proposition}[Noise term behavior]\label{prop:ntb}
    Let $\|\mathbf{b}\|^2 = o\left(\nu_1(d)\right)$, $\mathbf{b}^\top (\boldsymbol{\Sigma} + \boldsymbol{\Sigma}_{\psi(z)}) \mathbf{b} = o\left(\nu_2(d)\right)$, $\mathbf{b}^\top \boldsymbol{\Sigma}^{-1} \mathbf{b} = o\left(\nu_3(d)\right)$, $\mathbf{b}^\top (\boldsymbol{\Sigma}^{-1} + \boldsymbol{\Sigma}^{-1} \boldsymbol{\Sigma}_{\psi(\mathbf{Z})} \boldsymbol{\Sigma}^{-1}) \mathbf{b} = o\left(\nu_4(d)\right)$, and $\mathbf{b}^\top (\boldsymbol{\Sigma} + \boldsymbol{\Sigma}_{\psi(z)})^{-1} \mathbf{b} = o\left(\nu_5(d)\right)$. Here $\nu_i$ denotes the rates of growth with regard to $d$.

    Assume the distribution $P$ follows the structural causal model in Eq. \ref{eq:scm}, and the following conditions hold: \textbf{1.} $\lim_{d \to \infty} \frac{\nu_1(d)}{\nu_2(d)} \to \infty$, \textbf{2.} $\lim_{d \to \infty} \frac{\nu_3^2(d)}{\nu_4(d)} \to \infty$ and \textbf{3.} $\lim_{d \to \infty} \nu_5(d) \to \infty$.

    The following convergence properties hold: $\gamma^2(\mathbf{w}_S) \to \infty$ if condition \textbf{1} holds, $\gamma^2(\mathbf{w}_F) \to \infty$ if condition \textbf{2} holds, $\gamma^2(\mathbf{w}_D) \to \infty$ if condition \textbf{1}, \textbf{2} or \textbf{3} holds.


\end{proposition}

In general, the above conditions reflect the fact that $\mathbf{b}$ is unaligned with $\boldsymbol{\Sigma}$, meaning that large values of $\mathbf{b}$ correspond to small values of $\boldsymbol{\Sigma}$ and $\boldsymbol{\Sigma}_{\psi(Z)}$. This relationship can also be interpreted in terms of the growth of the largest eigenvalue of $\boldsymbol{\Sigma}$ or of $\|\mathbf{b}\|^2$, independently. All of these conditions are related to the observation that as the dimensionality increases, $Y^x$'s distribution contains 'more signal' relative to its noise level. This phenomenon occurs, for example, when the sources of noise are limited and the resolution of the observations is increased. We provide further details and insights on these assumptions in Sec. \ref{sec:noise_term_behavior}.
As discussed above, a strong SNR indicates that the recovered signal is closer to the information bottleneck $\phi(x)$. More importantly, it also implies better separability of the distributions of $Y^x$ along the projected axis. This can be formalised by considering the Fisher information of $\mathbf{w}^\top Y^x$ with respect to $x$, given by:
\begin{align*}
I_{\mathbf{w}}(x) = \mathbb{E} \left[U(x) U(x)^\top \right],
\end{align*}
with $U(x) =  \nabla_x \log P(\mathbf{w}^\top Y \mid  do(X= x)$ denoting the score function.
We now show that for linear models, the Fisher information and the SNR of $\mathbf{w}^\top Y^x$ are equivalent up to a positive scaling factor.
\begin{proposition}[Equivalence between Fisher information and SNR]\label{prop:SNR_FI_equiv}
    Consider a SCM as described in \eqref{eq:scm}, and let the intervention function be $\phi(x) = \mathbf{v}^\top x$, where $\mathbf{v} \in \mathbb{R}^d$. Then, the SNR is proportional to the Fisher Information of the intervention, i.e.   $I _{\mathbf{w}}(x) = \alpha \gamma^2(\mathbf{w})$ with $\alpha \in \mathbb{R}^+$.
\end{proposition}
Applying this result to $T_D$, we obtain an optimality guarantee in terms of Fisher information.
\begin{corollary}
    Under the assumptions of Prop. \ref{prop:SNR_FI_equiv}, the optimal solution $\mathbf{w}_D$ maximises the Fisher information $I_{\mathbf{w}}(x)$.
\end{corollary}
A similar result to Prop. \ref{prop:ntb} can also be derived for Fisher information under the assumption of a linear effect of $X$ on $Y$. Thus, the optimality conditions for recovering the bottleneck structure $\phi(x)$ translate into conditions for the discriminative power of the learned representation. 
In this setting, maximising the SNR is equivalent to maximising the Fisher information, which quantifies the sensitivity of the projected distribution to changes in the intervention parameter. This can be better understood by examining the relationship between Fisher information and the Kullback-Leibler divergence (see \ref{sec:IT_and_SNR}). Specifically, it measures the distance between parametric distributions, where in this case, the parameter corresponds to the intervention value. For linear models, $T_D$ is optimal as it maximises the distributional divergence induced by infinitesimal perturbations of the intervention. This enhances the discriminative power of the learned representation across different interventions. Moreover, higher Fisher information indicates that the learned representation retains more information about the intervention. 
%
\subsection{Testing the presence of a direct effect}\label{sec:lantent}
We now explore a direct implication of our problem formulation. Since we are maximising a test statistic for conditional independence testing, we can derive the distribution of the loss function under the null hypothesis. Consequently, one can reject the hypothesis of conditional independence at level $ \alpha $ if the value of the loss function, specifically the largest eigenvalue $ \lambda_{1} $, exceeds a critical threshold. 
%
\begin{proposition}[Distribution of $\lambda_F$ under conditional independence] \label{prop:lambda_F_distrib} 
    Let the distribution $P$ be induced by the SCM in \eqref{eq:scm} with linear assignments and Gaussian noise, and assume $p = q = 1$. Under the null hypothesis $H_0: X \indep Y \mid Z$, the largest root $\lambda_F$ is $F$-distributed such that $(dfn/dfd)\lambda_F\sim F(dfd, dfn)$ where $dfn = d$ and $dfd=n-p-r-1$.
\end{proposition}  

Finding the distribution of $\lambda_D$ is more challenging. Instead, we establish an upper bound on $\lambda_D$'s distribution, allowing the distribution of $\lambda_F$ to serve as a proxy for computing upper bounds on the p-values of $\lambda_D$.  

\begin{proposition}[Upper bound on $\Lambda_D$ under conditional independence]  
    Under similar assumptions as in Prop \ref{prop:lambda_F_distrib} we have under the null hypothesis $H_0: X \indep Y \mid Z$ that $P(\Lambda_D \geq \lambda_D | H_0) \leq P(\Lambda_F\geq\lambda_D |H_0)$.
\end{proposition}  

Testing is straightforward by rejecting the null hypothesis if $(dfd/dfn) \lambda_1$ deviates sufficiently from $F(dfn, dfd)$. This property is useful for testing whether the learned representation (Sec. \ref{sec:latent_structure_recovery}) captures a meaningful effect of $X$ on $Y$.

\section{Experiments}\label{sec:Experiments}

In this section, we present the results of our extensive simulation experiments designed to support our theoretical findings. Additionally, we provide a straightforward use case from climate science detection and attribution to illustrate the practical relevance of our approach. The code for all experiments is available at this \href{https://anonymous.4open.science/r/DEA_UAI2025-EBBF/}{github repository}.

\subsection{Simulation experiments}\label{sec:simulations}

\begin{figure}
    \centering
    \includegraphics[width=1\linewidth]{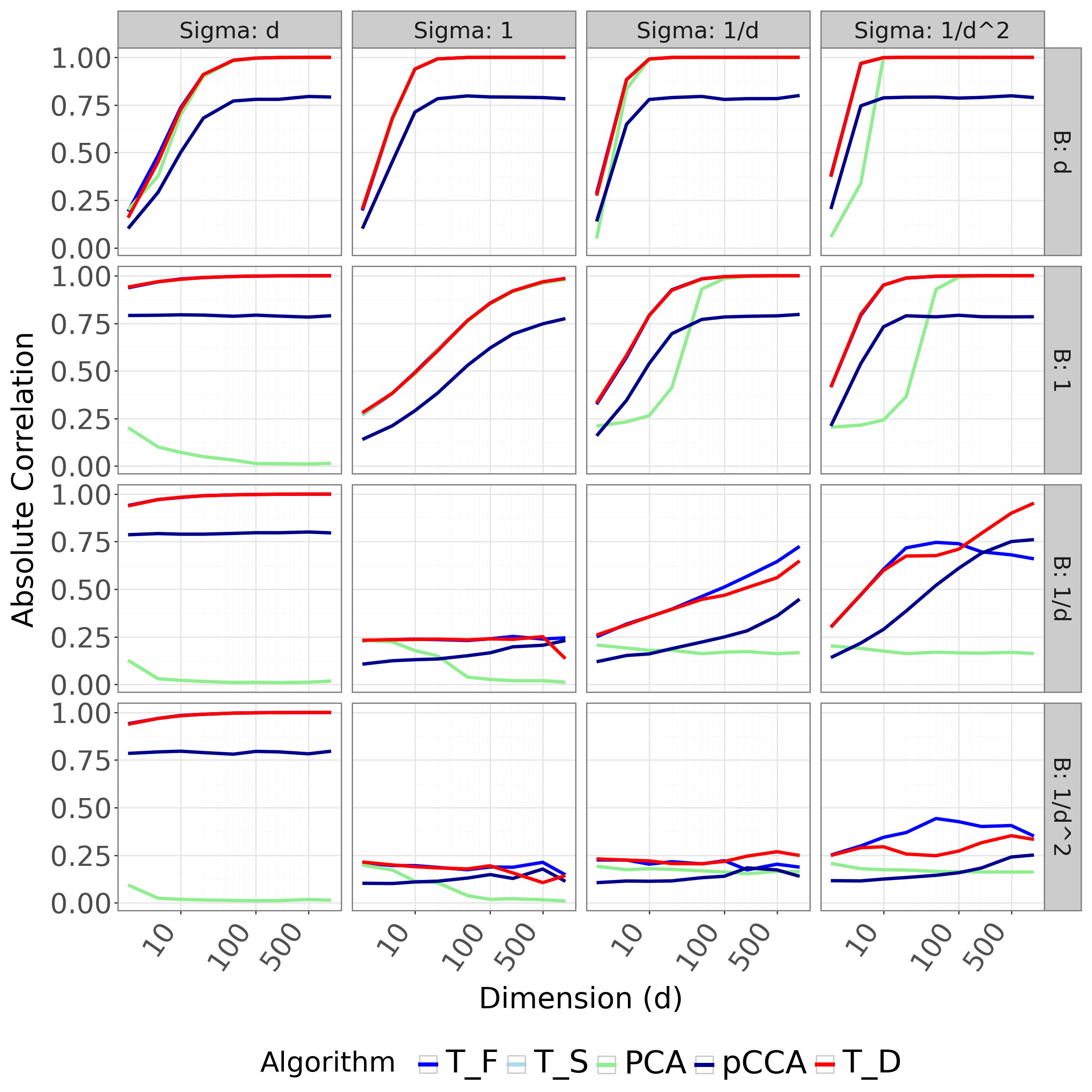}
    \caption{Correlation between $\mathbf{w}^\top Y$ and $\phi(X)$ as $d$ increases. $T_D$ consistently outperforms all methods, recovering $\phi(X)$ as $d$ grows, provided that $\mathbf{b}$ faster than $\mathbf{\Sigma}$. See Fig.~\ref{fig:DR_noise_behavior_Noise} for the (5, 95) percentiles.}
    \label{fig:DR_noise_behavior_noNoise}
\end{figure}

\begin{figure*}[ht]
    \centering
    \includegraphics[width=1\linewidth]{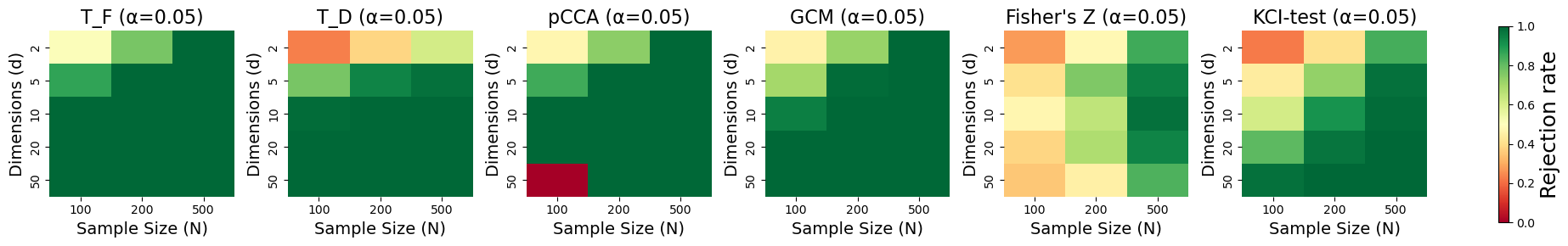}
    \caption{ Power of the test for $ \alpha =0.05$. A detailed experiments with different values $\alpha$ is available in Fig. \ref{fig:power_all}}
    \label{fig:test_results_linear}
\end{figure*}

We simulate data from a linear SCM with Gaussian noise, where $Z$ acts as a confounder for both $X$ and $Y$ \eqref{eq:scm}. The noise terms $N_x$ and $N_z$ are independent, with $f_a$ set to the identity for the linear case. For the nonlinear case, we define $f_a(z) := \exp(-z^2/2)\sin(az)$, where $a$ controls nonlinearity. The coefficients $\Gamma, \mathbf{b}, \mathbf{C}, \mathbf{D}$ are uniformly sampled from $[0, 1]$. We run 20 repetitions for each sample size $n$ and dimension $d$, reporting median values and quartiles.

\paragraph{Causal Effect Representation}
We assess the performance of our algorithm in recovering the direct effect of $X$ on $Y$, modeled as $f_a(\Gamma^\top X)$. The recovery is tested as $d$ increases and with varying noise structures. We set $p = r = 10$, and use $n = 4000$ samples for robust evaluation. Performance is evaluated by the absolute correlation between $\mathbf{w}^\top Y$ and $f_a(\Gamma^\top X)$, comparing nested models ($T_S$, $T_F$, $T_D$) against PCA and partial CCA (pCCA) as baselines.
To understand the contexts where learning algorithms may fail to fully recover $\phi(X)$, we consider various configurations of $\mathbf{\Sigma}$ and $\mathbf{b}$. We set $\mathbf{\Sigma}$ to be diagonal and explore four sets of entries for $Diag(\mathbf{\Sigma})$ and $\mathbf{b}$: $(1, \dots, i, \dots, d)$, $(1, \dots, 1)$, $(1, \dots, 1/i, \dots, 1/d)$, and $(1, \dots, 1/i^2, \dots, 1/d^2)$. Our main observation is that when $\mathbf{b}$ grows slowly relative to $\mathbf{\Sigma}$, none of the methods fully recover the signal. Specifically, pCCA tends to converge to a correlation of approximately 0.75, as it only recovers the part of $\phi(X)$ independent of $Z$—the signal correlated with $\psi(Z)$ is regressed out from both residuals before regression. This behavior is clearer in Appendix figure \ref{fig:DR_noise_behavior_indep}, where $X$ and $Z$ are simulated as independent variables, and pCCA can recover $\phi(X)$. Additionally, we observe that $T_F$ and $T_D$ outperform $T_S$ when $\mathbf{b}$ grows too slowly relative to the noise. Both $T_F$ and $T_D$ effectively control the variance contributions from $\mathbf{\Sigma}$, resulting in better performance in these challenging contexts (Appendix figure \ref{fig:DR_noise_behavior_noNoise}).
We analyse recovery across various noise configurations for $\mathbf{\Sigma}$: Diagonal, Full-rank, and Low-rank (rank = 10). We also test three weighting schemes: \textit{equal}, \textit{strong\_N\_Y}, and \textit{strong\_Z}, setting $(u, v, w)$ in \eqref{eq:simulation_SCM} by $(1/3, 1/3, 1/3)$, $(0.1, 0.1, 0.8)$, and $(0.1, 0.8, 0.1)$, respectively. As shown in Appendix figure \ref{fig:DR}, $T_D$ consistently outperforms other methods, with correlation approaching 1 as $d$ increases. Similar trends are observed in nonlinear and high-dimensional cases (Appendix figures \ref{fig:DR_nonlinear}, \ref{fig:DR_high_dimensional}).

\paragraph{Level and Power of the Test}
We assume that the data are generated from a linear SCM with Gaussian noise, where $f_a(Z) = Z$ and set $p = r = q = 1$. Our analysis compares tests based on the optimisation of $T_F$ and $T_D$ against four common conditional independence (CI) tests: partial CCA ~\citep{Rao1969}, the Generalised Covariance Measure (GCM)~\citep{Shah2018TheHO}, Fisher's Z test~\citep{kalisch2007estimating}, and the Kernel Conditional Independence (KCI) test~\citep{zhang2012}. The primary focus is on test performance with respect to sample size and $Y$'s dimensionality.
All tests maintain valid control of false positives when $d < n$ (see Appendix \ref{fig:type_I_control}), ensuring effective Type I error control. However, for test power, Fisher's Z and KCI show lower performance, especially for small samples and large $d$, due to their broader hypothesis set $\mathcal{P}$, which includes potentially nonlinear relationships. 
Tests based on $T_F$, $T_D$ and pCCA leverage $Y$'s dimensionality, showing better performance with higher dimensions for fixed sample sizes. This contrasts with Fisher's $Z$, which performance does not increases with $d$.

\subsection{Real-world experiments}

We present two real-world climate detection and attribution experiments: the first leverages the algorithm's ability to learn disentangled representations, and the second applies $T_D$ to test causal effects.

\paragraph{Separating internal climate variability from the externally forced response.}


We evaluate the ability of our method to disentangle internal climate variability from the externally forced response using temperature fields from CESM2 historical climate simulations~\citep{danabasoglu2020community}. Use of the optimal projection $\mathbf{w}_D$ is compared against two commonly used baselines in climate science: \textit{Detrending} and \textit{Dynamical Adjustment}~\citep{Sippel2019}. 
To achieve this, we model internal variability using Sea Level Pressure (SLP) as a proxy and estimate the externally forced response using a smoothed version of the Global Mean Temperature (GMT). $T_D$ learns a projection that isolates the internal component of temperature fluctuations while preserving their dynamical structure. Once trained, the model allows us to separate the forced and internal components of temperature fields.
Figure~\ref{fig:mse_boxplot} presents the mean squared error (MSE) for trend estimation across different algorithms. $T_D$ performs comparably to Detrending for reconstructing forced trends but performs better in recovering internal variability trends, providing better worst-case control. The spatial distribution of estimated internal trends (Figures~\ref{fig:trends_maps_DEA} and \ref{fig:trends_maps_Detrending}) further highlights that both methods capture large-scale patterns but tend to underestimate trends in polar regions. Additionally, Figure~\ref{fig:climate_experiment_forced_response_trends_TS} illustrates that  $T_D$ effectively reconstructs the forced response across different locations, although both $T_D$ and Detrending struggle in highly variable regions.
Overall, our approach provides a principled framework for disentangling forced and internal climate variability.
\paragraph{Climate change attribution.}

\begin{table}
    \centering
    \caption{Performance comparison of different approaches for detecting various effects. Bold values indicate the lowest Type II Error and Type I Error at level $5\%$.}
    \label{tab:effects}
    \begin{tabular}{llcc}
        \toprule
        \bfseries Effect & \bfseries Approach & \bfseries Type II Err. & \bfseries Type I Err. \\
        \midrule
        $\mathrm{CO}_2$ & DEA & $\mathbf{0.00}$ & $\mathbf{0.00}$ \\
                        & GMT Reg & $0.06$ & $0.30$ \\
                        & EOF & $0.06$ & $0.30$ \\
        \midrule
        $\mathrm{CH}_4$ & DEA & $\mathbf{0.52}$ & $\mathbf{0.00}$ \\
                        & GMT Reg & $0.70$ & $0.30$ \\
                        & EOF & $0.74$ & $0.26$ \\
        \midrule
        Aerosol & DEA & $\mathbf{0.00}$ & $\mathbf{0.04}$ \\
                & GMT Reg & $0.76$ & $0.24$ \\
                & EOF & $0.76$ & $0.24$ \\
        \midrule
        Land Use & DEA & $\mathbf{0.00}$ & $\mathbf{0.14}$ \\
                 & GMT Reg & $0.36$ & $0.64$ \\
                 & EOF & $0.74$ & $0.26$ \\
        \bottomrule
    \end{tabular}
\end{table}

In this experiment, we examine the direct effects of external forcing and investigate whether external forcing factors—such as aerosols, $\mathrm{CO}_2$, $\mathrm{CH}_4$, and land use have a direct effect on the annual mean temperature field ($Y_{\text{factual}}$). Using 50 historical climate simulations from CESM2, we compute counterfactual temperature fields ($Y_{\text{counterfactual}}$), following the methodology described in Eq. \ref{eq:intern_forced} in the supplementary materials. We apply the algorithm $T_D$ to test for the significance of each forcing ($X$) while controlling for the effects of the others ($Z$). Our results are compared to two common approaches in climate attribution~\citep{lean2008natural}: regression-based tests where forcings are assessed for their significance in predicting climate patterns, specifically Global Mean Temperature (spatial average) and the first Empirical Orthogonal Function (EOF) of the climate field. The findings demonstrate that our method effectively controls type I error (when applied to $Y_{\text{factual}}$) and type II error (when applied to $Y_{\text{counterfactual}}$) and outperforms the other approaches. These results highlight the potential of our method in attributing causal effects of external forcing, with implications for its use in analysing observational data, such as the ERA5 or HADCRUT datasets.

\section{Conclusion}

This paper proposes a novel framework for recovering the direct effect of low-rank interventions in multivariate response variables. Our approach combines conditional independence testing and causal representation learning, enabling robust estimation of direct causal effects in multivariate settings.  
We showed that the choice of test statistic \(T\) significantly influences algorithm performance, with different choices yielding varying effectiveness. Notably, the learning algorithm that controls noise variance exhibits stronger theoretical guarantees and improved performance in simulations, even in nonlinear settings.  
Our results highlight that performance depends on noise matrix assumptions, particularly as dimensionality \(d\) increases, leading to better discriminative power of intervention distributions. Furthermore, the loss function serves as a statistic in CI tests, allowing us to assess whether \(X\) significantly affects \(Y\) while enhancing interpretability. Our approach ensures robustness in multivariate settings and enables extensions to other CI test statistics and regression models, fostering broader applicability.  
Future work will derive the distribution of the optimal learning loss under null and alternative hypotheses to enhance test power. We will also explore nonlinear representations via projection into a reproducing kernel Hilbert space and assess cases where the effects of \(X\) and \(Z\) on \(Y\) are not linearly separable. Additionally, we aim to further investigate this problem from an information-geometric perspective.

\newpage

\bibliographystyle{abbrvnat} 
\bibliography{references2}  

\newpage

\onecolumn

\appendix

\title{Supplementary Material}
\maketitle

\appendix

\tableofcontents

\onecolumn

\section{Further methodological details}
We provide some further details and intuitions about our approach, starting by giving some insights it has with information theory.
\subsection{An information theory perspective on nested models test maximisation}\label{sec:IT_and_SNR}

In an information-theoretic framework~\citep{thomas2006elements}, the nested models residuals \eqref{eq:simple_population_loss} can be interpreted as measures of uncertainty—quantified by entropy—regarding \( Y \) given \( X \) and \( Z \). The proposed statistic then aims to maximise the conditional mutual information  
\[
I(X; \mathbf{w}^\top Y \mid Z) = H(\mathbf{w}^\top Y \mid Z) - H(\mathbf{w}^\top Y \mid Z, X),
\]
where \( H(\mathbf{w}^\top Y \mid Z) \) and \( H(\mathbf{w}^\top Y \mid Z, X) \) denote the corresponding conditional entropies. This aligns with the well-established connection between conditional mutual information, causality, and conditional independence~\citep{janzing2013quantifying}.  

Moreover, Proposition~\ref{prop:SNR_FI_equiv} establishes that, in the linear case with information bottleneck, maximising the signal-to-noise ratio (SNR) is equivalent to maximizing Fisher information. This implies that the proposed algorithm optimally distinguishes between the interventional distributions \( p(Y \mid do(X = x)) \) and \( p(Y \mid do(X = x + \delta x)) \), improving their separability under small interventions. This follows from the well established connection between Fisher information and the Kullback–Leibler divergence.

\begin{proposition}
Let \( P(Y \mid x) \) be a probability distribution over \( Y \) parameterised by \( x \in \mathbb{R}^d \). Consider a small perturbation \( \delta x \) such that \( P(Y \mid x + \delta x) \) remains close to \( P(Y \mid x) \). Then, the Kullback–Leibler divergence between these two distributions admits the following second-order expansion:
\[
D_{\mathrm{KL}}(P(Y \mid x) \,\|\, P(Y \mid x + \delta x))
= \frac{1}{2} \delta x^\top I(x) \delta x + O(\|\delta x\|^3),
\]
where \( I(x) \) is the \emph{Fisher information matrix}, given by
\[
I_{\mathbf{w}}(x) = \mathbb{E} \left[ U(x) U(x)^\top \right],
\]
with \( U(x) =  \nabla_x \log P(\mathbf{w}^\top Y \mid  X= x) \) denoting the score function.
\end{proposition}

The proof is provided in Appendix~\ref{subsec:FI_SNR}.  

This result formalises the intuition that our algorithm identifies a subspace that maximally separates distributions under infinitesimal intervention perturbations, enhancing their distinguishability.

\subsection{Partial correlation analysis as the maximisation of a conditional independence test statistic}\label{supp:pCCA}
We briefly outline how the partial Canonical Correlation Analysis (CCA) test, originally introduced by \citet{Rao1969}, can be interpreted within our framework. Specifically, we show that it can be viewed as the maximisation of a partial correlation test between $\mathbf{w}^\top Y$ and $X$ when adjusted for $Z$.

Let the population residuals after regressing out \( Z \) be defined as:
\begin{align*}
    R_x(\mathbf{v}) &= \mathbf{v}^\top X - \mathbb{E}[\mathbf{v}^\top X \mid Z], \\
    R_y(\mathbf{w}) &= \mathbf{w}^\top Y - \mathbb{E}[\mathbf{w}^\top Y \mid Z].
\end{align*}

Assuming a linear relationship between \( X \) and \( Y \), the conditional independence statistic can be expressed as:
\begin{align}\label{eq:population_loss_partialCCA}
    T_{\text{C}}(X, Y, Z; \mathbf{w}, \mathbf{v}) &= \text{artanh}(\text{corr}(R_x(\mathbf{v}), R_y(\mathbf{w}))).
\end{align}
Under the null hypothesis of conditional independence, and assuming that \( R_x \) and \( R_y \) are linearly related and follow a Gaussian distribution, it can be shown that \( T_C \) is asymptotically normally distributed. Since the \( \text{artanh} \) function is monotonic, maximising the CIT statistic is equivalent to maximizing the partial correlation test statistic.

It also share similarities to the statistic proposed in \citep{shah2018}, with the key difference being the normalisation used. While \citet{shah2018} proposed a statistic based on the covariance of the residuals normalised by the variance of their product, we normalise by the product of the variances of the residuals, resulting in a correlation coefficient. Although the approach in \citep{shah2018} is known to have power against alternatives under very weak assumptions—specifically, that the convergence rate of the estimators of the conditional expectations results in an error product rate of $o(n^{-1})$—it is less straightforward to derive explicit formulations for $\mathbf{v}$ and $\mathbf{w}$ from their method.

\paragraph{Empirical estimator for partial CCA} 

We are given two unbiased estimators $\hat{f}_x(Z)$ and $\hat{f}_y(Z)$ of respectively $\mathbb{E}[X|Z]$ and $\mathbb{E}[Y|Z]$. We denote by $ \hat{\mathbf{R}}_{x} $ and $ \hat{\mathbf{R}}_{y} $ the (empirical) residuals obtained from the predictions of $ \mathbf{X} $ and $ \mathbf{Y} $, respectively $\hat{\mathbf{R}}_{x} = \mathbf{X} - \hat{f}_x(\mathbf{Z})$ and $\hat{\mathbf{R}}_{y} = \mathbf{Y} - \hat{f}_y(\mathbf{Z})$. Similarly, the maximisation of the loss can be obtained via a generalised eigenvalue decomposition 
\begin{equation}\label{eq:pcorr_empirical_loss}
\hat{\mathbf{\Sigma}}_{\mathbf{R}_x\mathbf{R}_y}^\top \hat{\mathbf{\Sigma}}_{\mathbf{R}_y} ^{-1}\hat{\mathbf{\Sigma}}_{\mathbf{R}_x \mathbf{R}_y} \mathbf{w} = \lambda     \hat{\mathbf{\Sigma}}_{\mathbf{R}_y}\mathbf{w},
\end{equation}
where $\hat{\mathbf{\Sigma}}_{\mathbf{R}_x \mathbf{R}_y}$ is the sample covariance of $ \hat{\mathbf{R}}_{x} $ and $\hat{\mathbf{R}}_{y} $, and $\hat{\mathbf{\Sigma}}_{\mathbf{R}_y}$ is the sample covariance of $ \hat{\mathbf{R}}_{y} $.

\subsection{Empirical estimators}\label{sec:Estim_details}

We provide additional details regarding our estimators, focusing on three key aspects: the estimation of conditional expectations, the extraction of multiple components, and the stability of the solutions obtained through the Generalised Eigenvalue (GEV) problem.

\paragraph{Estimation of the Conditional Expectation} 
We have, so far, assumed the availability of unbiased estimators for the conditional expectations $\mathbb{E}[Y|X, Z]$ and $\mathbb{E}[Y|Z]$. In practice, these estimators should be selected based on domain-specific knowledge. In our case, we use the OLS estimator for the linear case and random forests for the nonlinear case.

\paragraph{Further Components}\label{sec:more_components}
Until now, we have considered the case where $ q = 1 $, assuming that the dimensionality of the direct effect of $ X $ on $ Y $ is rank one. In a manner analogous to the power iteration method \citep{Mises1929}, we can extract additional components by employing a deflation technique. We now provide an algorithm for this approach.

\begin{algorithm}
\caption{Power Method for DEA}
\label{algo:compact_GEV_power_iteration}

\KwIn{Matrices $ \mathbf{X} $, $ \mathbf{Y} $, $ \mathbf{Z} $, 
components $K$}
\KwOut{Matrix $ \mathbf{W} = [\mathbf{w}_1, \mathbf{w}_2, \dots, \mathbf{w}_K] $}
Initialise $ \mathbf{W} \gets [~~] $\\
Solve $ \mathbf{w}_1$ 
by maximising the empirical version of \eqref{eq:simple_population_loss}, \eqref{eq:population_loss_F} or \eqref{eq:population_loss_detect}\\
Normalise $\mathbf{w}_1 \gets \frac{\mathbf{w}_1}{\| \mathbf{w}_1 \|^2}$\\
Append $ \mathbf{w}_1 $ to $ \mathbf{W} $\\
\For{$k = 1$ \KwTo $K$}{
    Deflate $ \mathbf{Y}^{(k+1)} \gets \mathbf{Y}^{(k)} - \sum_{i=1}^{k} \mathbf{Y}^{(k)} \mathbf{w}_i \mathbf{w}_i^\top $\\
    Solve $ \mathbf{w}_k \gets \arg\max_{\mathbf{w}} \mathbf{T}_n(\mathbf{X}, \mathbf{Y}^{(k)} \mathbf{w}, \mathbf{Z}) $\\
    Normalise $ \mathbf{w}_k \gets \frac{\mathbf{w}_k}{\| \mathbf{w}_k \|} $
    Append $ \mathbf{w}_k $ to $ \mathbf{W} $
}
\Return $\mathbf{W}$
\end{algorithm}

\paragraph{Stability of the Solution}\label{sec:stability}
The stability of the solutions is influenced by the covariance matrices $ \hat{\boldsymbol{\Sigma}}_{\text{full}} $ and $ \hat{\boldsymbol{\Sigma}}_{R_y} $, which may be ill-conditioned due to the characteristics of the noise term $ N_Y $. This can complicate the GEV resolution. To mitigate this issue, we use a regularisation strategy that modifies the covariance matrices as $\hat{\boldsymbol{\Sigma}}_{\text{full}} + \delta \mathbf{I}, \quad \text{and} \quad \hat{\boldsymbol{\Sigma}}_{R_y} + \delta \mathbf{I}$, where $ \delta $ is a small constant (typically $ 10^{-8} $) that stabilises the smallest eigenvalues. 

Optimising the regularisation parameter more effectively might be crucial in the context of high-dimensional response variables. A promising approach could be the Ledoit-Wolf regularisation strategy, as proposed by \cite{ledoit2004well}.

\subsection{Noise term behavior}\label{sec:noise_term_behavior}

The conditions outlined in Proposition \ref{prop:ntb} may initially seem complex, so we provide a more intuitive explanation here. In many practical scenarios, improving the signal-to-noise ratio becomes crucial as the dimensionality of $Y$ increases, which can occur when enhancing image resolution or adding sensors. Higher-dimensional data provides a richer representation of the system, enabling better separation of signal and noise and improving inference accuracy.

\paragraph{Example.} In climate science, we analyze global temperature patterns using climate observations. Let $Y$ represent the observed temperature field and $N_Y$ represent the observational errors arising from sensor limitations or model imperfections. The function $\phi(X)$ may capture internal climate variability (e.g., El Niño), while $\psi(Z)$ represents external forcing (e.g., greenhouse gas emissions). Increasing data granularity—through higher-resolution climate models, more sensors, or longer historical records—enhances the detectability of systematic climate responses while averaging out transient noise. As a result, the signal-to-noise ratio improves, making it easier to discern causal relationships and understand climate drivers.

The key insight is that algorithm performance depends on the structure of $\mathbf{b}$ and its interaction with noise terms. Performance improves when signal variance increases ($\|\mathbf{b}\|^2$ grows with $d$) in directions where noise covariances $\mathbf{\Sigma}$ and $\mathbf{\Sigma}_{\psi(Z)}$ are small. We can for example think about the simple case where the eigenvalue of the covariance matrix $\mathbf{\Sigma}_N$ decay quadractically as $d$ increases and where $\mathbf{b}=(1, \dots, 1)$. The estimator $T_D$ is optimal in that it maximises the signal-to-noise ratio under mild conditions (see Proposition \ref{prop:snr_max}). Convergence issues arise only in rare cases where $\mathbf{b}$, $\mathbf{\Sigma}$, and $\mathbf{\Sigma}_{\psi(Z)}$ decay at similar rates, as illustrated in Figure \ref{fig:DR_noise_behavior_noNoise}.

While these guarantees hold in the idealised population setting with infinite data, real-world applications often involve limited samples. In such cases, the theoretical insights may not directly translate into robust performance, necessitating regularisation techniques to prevent overfitting and improve estimation reliability in small datasets.

\section{Proofs}\label{sec:Proofs}

We now provide proof of our main theoretical results. As this will be useful for most of the theoretical development, we first get a result for the first eigenvector of each optimisation problem.

\subsection{Auxiliary lemma}

\begin{lemma}\label{lemma:EGV_sol}
Let $\mathbf{w}_S$, $\mathbf{w}_F$, and $\mathbf{w}_D$ denote the first eigenvectors associated with the optimisation problems in Eq. \eqref{eq:simple_population_loss}, Eq. \eqref{eq:population_loss_F}, and Eq. \eqref{eq:population_loss_detect}, respectively. The following properties hold:

\begin{enumerate}
    \item The eigenvector $\mathbf{w}_S$ is proportional $\mathbf{b}$, i.e., $\mathbf{w}_S \propto \mathbf{b}$, when maximizing Eq. \eqref{eq:simple_population_loss}. In the linear case it corresponds to the Gradient EDE.
    \item The eigenvector $\mathbf{w}_F$ is proportional to the direction of the inverse covariance-weighted true causal effect, i.e., $\mathbf{w}_F \propto \mathbf{\Sigma}^{-1} \mathbf{b}$, when maximizing Eq. \eqref{eq:population_loss_F}.
    \item The eigenvector $\mathbf{w}_D$ is proportional to the inverse of the sum of the covariances of the noise and confounding variables, i.e., $\mathbf{w}_D \propto (\mathbf{\Sigma} + \mathbf{\Sigma}_{\psi(Z)})^{-1} \mathbf{b}$, when maximizing Eq. \eqref{eq:population_loss_detect}.
\end{enumerate}
\end{lemma}

\begin{proof}
    Recall the definitions:
    \begin{align*}
        R_{\text{full}}^2(\mathbf{w}) &= \mathbb{E}[(\mathbf{w}^\top Y - \mathbb{E}[\mathbf{w}^\top Y | X, Z])^2], \\
        R_{\text{res}}^2(\mathbf{w}) &= \mathbb{E}[(\mathbf{w}^\top Y - \mathbb{E}[\mathbf{w}^\top Y | Z])^2], \\
        R_{\text{noise}}^2(\mathbf{w}) &= \mathbb{E}[(\mathbf{w}^\top Y - \mathbb{E}[\mathbf{w}^\top Y \mid X, Z=0])^2].
    \end{align*}

    From the model $Y^x = \mathbf{b}\phi(x) + \psi(Z) + N_y$, we derive:
    \begin{align*}
        R_{\text{full}}^2(\mathbf{w}) &= \mathbf{w}^\top \mathbf{\Sigma} \mathbf{w}, \\
        R_{\text{res}}^2(\mathbf{w}) &= \mathbf{w}^\top \mathbf{\Sigma} \mathbf{w} + \phi(x)^2 \mathbf{w}^\top \mathbf{b} \mathbf{b}^\top \mathbf{w}, \\
        R_{\text{noise}}^2(\mathbf{w}) &= \mathbf{w}^\top \mathbf{\Sigma} \mathbf{w} + \mathbf{w}^\top \mathbf{\Sigma}_{\psi(Z)} \mathbf{w}.
    \end{align*}

    The difference between the residual and full terms is:
    \[
        R_{\text{res}}^2(\mathbf{w}) - R_{\text{full}}^2(\mathbf{w}) = \phi(x)^2 \mathbf{w}^\top \mathbf{b} \mathbf{b}^\top \mathbf{w}.
    \]

    Each optimisation problem for $T_S$, $T_F$, and $T_D$ corresponds to a generalised eigenvalue problem of the form $\mathbf{N}^{-1}\mathbf{M}$, where $\mathbf{M} = \phi(x)^2 \mathbf{b} \mathbf{b}^\top$ is rank-1. Therefore, the first eigenvector $\mathbf{w}_1$ is proportional to $\mathbf{N}^{-1} \mathbf{b}$. The optimal solutions are:
    \begin{enumerate}
        \item $\mathbf{w}_S \propto \mathbf{b}$ for Eq. \eqref{eq:simple_population_loss},
        \item $\mathbf{w}_F \propto \mathbf{\Sigma}^{-1} \mathbf{b}$ for Eq. \eqref{eq:population_loss_F},
        \item $\mathbf{w}_D \propto (\mathbf{\Sigma} + \mathbf{\Sigma}_{\psi(Z)})^{-1} \mathbf{b}$ for Eq. \eqref{eq:population_loss_detect}.
    \end{enumerate}
\end{proof}

\subsection{Signal-to-Noise optimality }

\begin{proposition}[General optimality]
    Assuming $P$ is entailed in the SCM in Eq. \eqref{eq:scm}, we have that $\mathbf{w}_D$ is optimal.
\end{proposition}
\begin{proof}
    Recall that the optimal detector loss is defined as:
    \begin{align*}
        T_D &= \frac{R_{\text{res}}^2(\mathbf{w}) - R_{\text{full}}^2(\mathbf{w})}{R_{\text{noise}}^2(\mathbf{w})} = \frac{\phi(x)^2 \mathbf{w}^\top \mathbf{b} \mathbf{b}^\top \mathbf{w}}{\mathbf{w}^\top \mathbf{\Sigma} \mathbf{w} + \mathbf{w}^\top \mathbf{\Sigma}_{\psi(Z)} \mathbf{w}}.
    \end{align*}

    Using the results from \autoref{lemma:EGV_sol}, the signal-to-noise ratio is given by:
    \begin{align*}
        \gamma^2(\mathbf{w}) &= \frac{(\mathbf{w}^\top S(x))^2}{\mathbf{w}^\top \mathbf{\Sigma}_N \mathbf{w}} = \phi(x)^2 \frac{\mathbf{w}^\top \mathbf{b} \mathbf{b}^\top \mathbf{w}}{\mathbf{w}^\top (\mathbf{\Sigma} + \mathbf{\Sigma}_{\psi(Z)}) \mathbf{w}},
    \end{align*}
    where we note that $\psi(Z)$ and $N_y$ are assumed to be independent.

    This completes the proof.
\end{proof}

\begin{proposition}[Optimality under isotropic noise]
    Assuming that $P$ is entailed in the SCM in Eq. \eqref{eq:scm} and that $\mathbf{\Sigma}_N$ is isotropic, we have that both $\mathbf{w}_S$ and $\mathbf{w}_D$ are optimal. Moreover, if $\mathbf{\Sigma}_y$ is also isotropic, then $\mathbf{w}_F$ is also optimal.
\end{proposition}

\begin{proof}
    The proof follows straightforwardly from the assumption that $\mathbf{\Sigma}_N = \mathbf{\Sigma}_{\phi(z)} + \mathbf{\Sigma}$ is isotropic. Under this assumption, the constraint $\mathbf{w}^\top (\mathbf{\Sigma}_{\phi(z)} + \mathbf{\Sigma}) \mathbf{w}$ is equivalent to the constraint $\|\mathbf{w}\| = 1$. Therefore, the loss function $T_D$ simplifies to the form of $T_S$. As a consequence, the optimality of $\mathbf{w}_D$ stated in Prop. \ref{prop:snr_max} implies the optimality of $\mathbf{w}_S$.

    Similarly, when both $\mathbf{\Sigma}_N$ and $\mathbf{\Sigma}_y$ are isotropic, we observe that $T_F$ becomes equivalent to $T_S$. Since it has been established that if $\mathbf{\Sigma}_N$ is isotropic, $\mathbf{w}_S$ is optimal, we conclude that $\mathbf{w}_F$ is also optimal.
\end{proof}

\subsection{Noise term behavior}

\begin{proposition}[Noise Term Behavior]
    Let $\|\mathbf{b}\|^2 = o\left(\nu_1(d)\right)$, $\mathbf{b}^\top (\boldsymbol{\Sigma} + \boldsymbol{\Sigma}_{\psi(z)}) \mathbf{b} = o\left(\nu_2(d)\right)$, $\mathbf{b}^\top \boldsymbol{\Sigma}^{-1} \mathbf{b} = o\left(\nu_3(d)\right)$, $\mathbf{b}^\top (\boldsymbol{\Sigma}^{-1} + \boldsymbol{\Sigma}^{-1} \boldsymbol{\Sigma}_{\psi(\mathbf{Z})} \boldsymbol{\Sigma}^{-1}) \mathbf{b} = o\left(\nu_4(d)\right)$, and $\mathbf{b}^\top (\boldsymbol{\Sigma} + \boldsymbol{\Sigma}_{\psi(z)})^{-1} \mathbf{b} = o\left(\nu_5(d)\right)$.

    Assume that $\phi(X) \in \mathbb{R}$, the distribution $P$ follows the structural causal model in \ref{eq:scm}, and the following conditions hold:
    \begin{enumerate}
        \item $\lim_{d \to \infty} \frac{\nu_1(d)}{\nu_2(d)} \to \infty$,

        \item $\lim_{d \to \infty} \frac{\nu_3^2(d)}{\nu_4(d)} \to \infty$,

        \item $\lim_{d \to \infty} \nu_5(d) \to \infty$.
    \end{enumerate}

    Under these conditions, the following convergence properties hold:
    \begin{enumerate}
        \item $\gamma^2(\mathbf{w}_S) \to \infty$ if condition 1) holds,

        \item $\gamma^2(\mathbf{w}_F) \to \infty$ if condition 2) holds,

        \item $\gamma^2(\mathbf{w}_D) \to \infty$ if condition 1), 2), or 3) holds.
    \end{enumerate}
\end{proposition}

\begin{proof}
    Recalling that we have
    \begin{align}
        R_{\text{full}}^2(\mathbf{w}) &= \mathbf{w}^\top \mathbf{\Sigma}\mathbf{w}\\
        R_{\text{res}}^2(\mathbf{w}) - R_{\text{full}}^2(\mathbf{w}) &= \phi(x)^2  \mathbf{w}^\top \mathbf{b}\mathbf{b}^\top  \mathbf{w}\\
        R^2_{\text{noise}} &= \mathbf{w}^\top \mathbf{\Sigma}\mathbf{w} + \mathbf{w}^\top \mathbf{\Sigma}_{\psi(Z)} \mathbf{w}
    \end{align}

    Substituting these into the signal-to-noise ratio in Eq. \eqref{eq:SNR}, we obtain:
    \begin{align}
        \gamma^2(\mathbf{w}_S) &= \frac{\|\mathbf{b}\|^2 \phi(x)^2}{\mathbf{b}^\top\mathbf{\Sigma}\mathbf{b} + \mathbf{b}^\top\mathbf{\Sigma}_{\psi(Z)}\mathbf{b}}, \\
        \gamma^2(\mathbf{w}_F) &= \frac{(\mathbf{b}^\top\mathbf{\Sigma}^{-1}\mathbf{b})^2 \phi(x)^2}{\mathbf{b}^\top\mathbf{\Sigma}^{-1}\mathbf{b} + \mathbf{b}^\top \mathbf{\Sigma}^{-1}\mathbf{\Sigma}_{\psi(Z)} \mathbf{\Sigma}^{-1}\mathbf{b}}, \\
        \gamma^2(\mathbf{w}_D) &= \mathbf{b}^\top(\mathbf{\Sigma} + \mathbf{\Sigma}_{\psi(Z)})^{-1}\mathbf{b} \phi(x)^2.
    \end{align}

    The convergence properties of the signal-to-noise ratio follow directly from these formulations, assuming $\phi(x)$ is bounded.
    Since \( \mathbf{w}_D \) maximises the SNR, if either \( \mathbf{w}_S \) or \( \mathbf{w}_F \) has an SNR that grows to infinity, then the SNR of \( \mathbf{w}_D \) will also tend to infinity.
\end{proof}

\subsection{Equivalence of Signal-to-Noise ratio and Fisher information}\label{subsec:FI_SNR}

\begin{proposition}[Equivalence between Fisher Information and SNR]
    Consider a SCM as described in \eqref{eq:scm}, and let the intervention function be $\phi(x) = \mathbf{v}^\top x$, where $\mathbf{v} \in \mathbb{R}^d$. Then, the SNR is proportional to the Fisher Information of the intervention, i.e.   $I _{\mathbf{w}}(x) = \alpha \gamma^2(\mathbf{w})$ with $\alpha \in \mathbb{R}^+$.
\end{proposition}

\begin{proof}
    Let $\mathbf{w}^\top Y^x \sim \mathcal{N}(\mathbf{w}^\top \mu(x), \mathbf{w}^\top \mathbf{\Sigma}_N \mathbf{w})$ denote the distribution of $\mathbf{w}^\top Y^x$.

    The log-likelihood for the intervention is given by:
    \begin{align*}
        \log p(Y \mid do(X=x)) &= C - \frac{1}{2} \mathbf{w}^\top (Y - \mu(x))^\top (\mathbf{w}^\top (\mathbf{\Sigma}_{\psi(z)} + \mathbf{\Sigma}) \mathbf{w})^{-1} (Y - \mu(x)) \mathbf{w},
    \end{align*}
    where $C$ is a constant relative to $x$. 

    The \textit{informant} $U(x)$ is the derivative of the log-likelihood with respect to $x$:
    \begin{align*}
        \frac{\partial}{\partial x} \log p(Y \mid do(X=x)) &= \frac{1}{2} \mu'(x)^\top (\mathbf{w}^\top \mathbf{\Sigma} \mathbf{w})^{-1} (Y - \mu(x)) \mathbf{w}.
    \end{align*}

    The Fisher information $I_\mathbf{w}(x)$ is the variance of the informant. Since the informant at the maximum likelihood has mean zero \citep[see][section 6]{lehmann2006theory}, we write:
    \begin{align*}
        I_\mathbf{w}(x) &= \mathbb{E}[U(x) U(x)^\top] \\
        &= \mathbb{E} \left[ \mu'(x)^\top (\mathbf{w}^\top \mathbf{\Sigma} \mathbf{w})^{-1} (Y - \mu(x)) \mathbf{w} \mathbf{w}^\top (Y - \mu(x))^\top (\mathbf{w}^\top \mathbf{\Sigma} \mathbf{w})^{-1} \mu'(x) \right].
    \end{align*}
    Using the fact that $\mathbb{E}[(Y - \mu(x))(Y - \mu(x))^\top] = \mathbf{\Sigma}$, we obtain:
    \begin{align*}
        I_\mathbf{w}(x) &= \mathbf{w}^\top \mu'(x) (\mathbf{w}^\top \mathbf{\Sigma} \mathbf{w})^{-1} \mu'(x) \mathbf{w}.
    \end{align*}

    Since $\mu(x) = \mathbf{b} \mathbf{v}^\top x$, we have $\mu'(x) = \mathbf{b} \mathbf{v}$. Additionally, $\mathbf{\Sigma} = \mathbf{\Sigma}_{\psi(z)} + \mathbf{\Sigma}$. Substituting these expressions into the Fisher information formula, we get:
    \begin{align*}
        I_\mathbf{w}(x) &= \mathbf{w}^\top \mathbf{b} \mathbf{v}^\top (\mathbf{w}^\top (\mathbf{\Sigma}_{\psi(z)} + \mathbf{\Sigma}) \mathbf{w})^{-1} \mathbf{v} \mathbf{b}^\top \mathbf{w} \\
        &= \frac{\mathbf{w}^\top \mathbf{b} \mathbf{v}^\top \mathbf{v} \mathbf{b}^\top \mathbf{w}}{\mathbf{w}^\top (\mathbf{\Sigma}_{\psi(z)} + \mathbf{\Sigma}) \mathbf{w}} \\
        &= \frac{\|\mathbf{v}\|_2^2}{\phi(x)^2} \gamma^2(\mathbf{w}).
    \end{align*}
\end{proof}

\begin{proposition}
Let \( P(Y \mid x) \) be a probability distribution over \( Y \) parameterised by \( x \in \mathbb{R}^d \). Consider a small perturbation \( \delta x \) such that \( P(Y \mid x + \delta x) \) remains close to \( P(Y \mid x) \). Then, the Kullback–Leibler divergence between these two distributions admits the following second-order expansion:
\[
D_{\mathrm{KL}}(P(Y \mid x) \,\|\, P(Y \mid x + \delta x))
= \frac{1}{2} \delta x^\top I(x) \delta x + O(\|\delta x\|^3),
\]
where \( I(x) \) is the \emph{Fisher information matrix}, given by:
\begin{align*}
    I_{\mathbf{w}}(x) = \mathbb{E} \left[ U(x) U(x)^\top \right].
\end{align*}
With $U(x) =  \nabla_x \log P(\mathbf{w}^\top Y \mid  X= x)$ the informant (or score) function.
\end{proposition}

\begin{proof}[Proof sketch]
   Assuming \( P(Y \mid x) \) is smooth in \( x \), we approximate it to its second order Taylor expansion:
   \[
   \log P(Y \mid x + \delta x) = \log P(Y \mid x) + \delta x^\top \nabla_x \log P(Y \mid x) + \frac{1}{2} \delta x^\top \nabla_x^2 \log P(Y \mid x) \delta x + O(\|\delta x\|^3).
   \]

   The KL divergence is defined as:
   \[
   D_{\mathrm{KL}}(P(Y \mid x) \,\|\, P(Y \mid x + \delta x))
   = \mathbb{E}_{Y \sim P(Y \mid x)} \left[ \log \frac{P(Y \mid x)}{P(Y \mid x + \delta x)} \right].
   \]

   Substituting into the KL divergence and using the property that \( \mathbb{E}_{Y \sim P(Y \mid x)} [\nabla_x \log P(Y \mid x)] = 0 \), the first-order term vanishes \citep[see][section 6]{lehmann2006theory}, leaving:
   \[
   D_{\mathrm{KL}}(P(Y \mid x) \,\|\, P(Y \mid x + \delta x))
   = -\frac{1}{2} \mathbb{E}_{Y \sim P(Y \mid x)} \left[ \delta x^\top \nabla_x^2 \log P(Y \mid x) \delta x \right] + O(\|\delta x\|^3).
   \]
   Since the Fisher information matrix is defined as \( I(x) = -\mathbb{E}[\nabla_x^2 \log P(Y \mid x)] \), we obtain:
   \[
   D_{\mathrm{KL}}(P(Y \mid x) \,\|\, P(Y \mid x + \delta x))
   = \frac{1}{2} \delta x^\top I(x) \delta x + O(\|\delta x\|^3).
   \]
\end{proof}

\subsection{Distribution of leading eigenvalues under conditional independence hypothesis}

\begin{proposition}[Distribution of $\lambda_F$ under conditional independence]
    Let the distribution $P$ be induced by the SCM in \eqref{eq:scm} with linear assignments and Gaussian noise, and assume $p = q = 1$. Under the null hypothesis $H_0: X \indep Y \mid Z$, the largest root $\lambda_F$ is $F$-distributed such that $(dfn/dfd)\lambda_F\sim F(dfd, dfn)$ where $dfn = d$ and $dfd=n-p-r-1$.
\end{proposition} 

\begin{proof}[Proof sketch]
    It can easily be shown that $\hat{R}^2_{\text{full}}$ and $\hat{R}_{\text{res}}^2$ follows $\chi^2$ distributions of respectively $d(n - p - r -1)$ and $d(n - p - r -1)$ degrees of freedom as they are computed as sums of squared Gaussian distributions. Their ratio can thus be shown to follow an F distribution with degrees of freedom $dfn=d$ and $dfd=n- p-r-1$.
    As the weights related to $Z$ are frozen when getting $\hat{R}^2_{\text{noise}}$, we have that it follows a $\chi^2$ with $d(n  - p - 1)$ degrees of freedom. Thus $\Lambda_D \sim F(p, n-p-1)$.
\end{proof}

We refer reader to the distribution of Roy's largest root, the Chow test \citep{chow1960tests} or the generalised linear hypothesis test (see e.g. \cite{anderson1958} chapter 7) as similar problems have been widely studied in the multivariate statistics literature \citep{anderson2003introduction, Bilodeau1999TheoryOM}.

\begin{proposition}[Upper Bound on $\Lambda_D$ Under Conditional Independence]  
    Under similar assumptions as in Prop \ref{prop:lambda_F_distrib} we have under the null hypothesis $H_0: X \indep Y \mid Z$ that $P(\Lambda_D \geq \lambda_D | H_0) \leq P(\Lambda_F\geq\lambda_D |H_0)$.
\end{proposition} 

\begin{proof}[Proof sketch]
As the conditioning set in the computation of root squared errors $R^2_{\text{noise}}$ is larger than of $R^2_{\text{full}}$, the empirical residuals $R^2_{\text{noise}}$ are always larger than $R^2_{\text{full}}$ thus we have that 
\begin{align*}
    \frac{R^2_{\text{res}} - R^2_{\text{full}}}{R^2_{\text{noise}}} \leq  \frac{R^2_{\text{res}} - R^2_{\text{full}}}{R^2_{\text{full}}}.
\end{align*}
Hence, we have that $P(\Lambda_D \geq \lambda_D | H_0) \leq P(\Lambda_F \geq \lambda_D|H_0)$ with $H_0: X\indep Y |Z$.
\end{proof}

Note that by using this upper bound we tend to lose power in the test procedure but we still control type I errors (we reject less than we would optimally do) and thus the test is valid. Further research should aim at discovering a better approximation for the distribution $\Lambda_D$.

\subsection{Convergence rates of $\mathbf{w}_S$, $\mathbf{w}_F$ and $\mathbf{w}_D$}
We first introduce an important theorem that will be useful for the proof.

\begin{theorem}[Davis-Kahan theorem \citep{davis1970rotation}]
Let $ \lambda^{(1)} - \lambda^{(2)} = \delta > 0 $ where $ \lambda^{(1)} > \lambda^{(2)} \geq \dots \geq \lambda^{(d)} $ be the eigenvalues of $ \mathbf{\Sigma} $ and $ \hat{\lambda}^{(1)} - \hat{\lambda}^{(2)} = \hat{\delta} > 0 $ where $ \hat{\lambda}^{(1)} > \hat{\lambda}^{(2)} \geq \dots \geq \hat{\lambda}^{(d)} $ be the eigenvalues of $ \mathbf{\hat{\Sigma}} $ and let $\mathbf{W}$ and $\mathbf{\hat{W}}$ their corresponding eigenvectors. We have that 
\begin{equation}
    \|\sin \Theta(\mathbf{W}, \mathbf{\hat{W}})\|_{op} \leq \frac{\|\mathbf{\Sigma - \mathbf{\hat{\Sigma}}\|_{op}}}{\max_j(|\hat{\lambda}_{j-1} - \lambda_j|, |\hat{\lambda}_{j+1} - \lambda_j|)}
\end{equation}
where $\Theta$ is a distance between subspaces. Similarly, for any $j$ we have that $\|\mathbf{\hat{w}}_j - \mathbf{w}_j\| \leq \sqrt{2}\sin \Theta(\mathbf{w}_j, \mathbf{\hat{w}_j})$.
    
\end{theorem}

We show that under common assumptions, specifically that there are two unbiased estimators $ \hat{g}_{\text{full}} $ and $ \hat{g}_{\text{res}} $ with convergence rates $ \kappa_1(n) $ and $ \kappa_2(n) $, the estimators proposed in Eq. \eqref{eq:simple_population_loss} are consistent with their population counterparts. Furthermore, we demonstrate that their convergence rate typically depends on the convergence rates $ \kappa_1(n) $ and $ \kappa_2(n) $.

\begin{proposition}[Convergence Rate of F-Test Based Losses]\label{prop:cv}
    Assume the following conditions hold:
 \begin{enumerate}
        \item $\mathbb{E} \left\| \hat{g}_{\text{full}}(\mathbf{X}_i, \mathbf{Z}_i) - \mathbb{E}[\mathbf{Y}_i \mid \mathbf{X}_i, \mathbf{Z}_i] \right\|^2 = o_P(\kappa_1(n))$,
        \item $\mathbb{E} \left\| \hat{g}_{\text{res}}(\mathbf{Z}_i) - \mathbb{E}[\mathbf{Y}_i \mid \mathbf{Z}_i] \right\|^2 = o_P(\kappa_2(n))$,
        \item $\lambda^{M}_{1} - \lambda^{M}_{2} = \delta_M > 0$, where $\lambda^{M}_{1} > \lambda^{M}_{2} \geq \dots \geq \lambda^{M}_{d}$ are the eigenvalues of $\mathbf{M}$,
        \item $\lambda^{N}_{1} - \lambda^{N}_{2} = \delta_N > 0$, where $\lambda^{N}_{1} > \lambda^{N}_{2} \geq \dots \geq \lambda^{N}_{d}$ are the eigenvalues of $\mathbf{N}$,
        \item $\mathbb{E} \left\| Y - \mathbb{E}[Y \mid X, Z] \right\|^2 \leq N_{\text{full}}$ and $\mathbb{E} \left\| Y - \mathbb{E}[Y \mid Z] \right\|^2 \leq N_{\text{res}}$.
    \end{enumerate}

    Let $\mathbf{w}_1$ be the optimal solution to Eq. \eqref{eq:simple_population_loss}, Eq. \eqref{eq:population_loss_F}, or Eq. \eqref{eq:population_loss_detect}, and let $\hat{\mathbf{w}}$ be the empirical solution to their respective empirical estimators. Under the given conditions, we have the following convergence result:
    \begin{align}
        \mathbb{E} \left[\|\mathbf{w}_1 - \hat{\mathbf{w}}\|_2^2 \right] = o \left( \sqrt{\kappa_1(n)} + \sqrt{\kappa_2(n)} \right).
    \end{align}
\end{proposition}

\begin{proof}
    Similar to what was done for the empirical estimators, the population loss can be written as an eigenvalue decomposition problem $\mathbf{N}^{-1}\mathbf{M}$, where $\mathbf{M} = \mathbf{\Sigma}_{\text{res}} - \mathbf{\Sigma}_{\text{full}}$, and $\mathbf{N}$ depends on the loss used. For simplicity, we consider $\mathbf{N} = \mathbf{I}$, which leads to the convergence result for the simple loss in Eq. \eqref{eq:simple_population_loss}. A similar reasoning can be applied to the convergence of the two other losses.

    Let us first decompose $\mathbf{\hat{\Sigma}_{\text{res}}}$ as follows:
    \begin{align}
        \mathbf{\hat{\Sigma}_{\text{res}}} &= \frac{1}{n}\sum_{i=1}^n (\mathbf{Y_i} - \hat{g}_{res}(\mathbf{Z}_i))(\mathbf{Y_i} - \hat{g}_{res}(\mathbf{Z}_i))^\top\\
        &= \frac{1}{n}\sum_{i=1}^n \left(\mathbf{Y_i} - g_{res}(\mathbf{Z}_i) + g_{res}(\mathbf{Z}_i) - \hat{g}_{res}(\mathbf{Z}_i)\right)\left(\mathbf{Y_i} - g_{res}(\mathbf{Z}_i) + g_{res}(\mathbf{Z}_i) - \hat{g}_{res}(\mathbf{Z}_i)\right)^\top\\
        &= \frac{1}{n}\sum_{i=1}^n \mathbf{N}_{y,i}^\top \mathbf{N}_{y,i} + \frac{2}{n}\sum_{i=1}^n \mathbf{N}_{y,i}^\top \left(g_{res}(\mathbf{Z}_i) - \hat{g}_{res}(\mathbf{Z}_i)\right) + \frac{1}{n}\sum_{i=1}^n \left(g_{res}(\mathbf{Z}_i) - \hat{g}_{res}(\mathbf{Z}_i)\right)^\top \left(g_{res}(\mathbf{Z}_i) - \hat{g}_{res}(\mathbf{Z}_i)\right)
    \end{align}
    where $\mathbf{N}_{y,i}$ is the population residual (noise) of sample $i$.

    We now aim to bound $\|\mathbf{\Sigma} - \mathbf{\hat{\Sigma}}\|_F$. Using the previous notation, we have:
    \begin{align}
        \|\mathbf{\Sigma} - \mathbf{\hat{\Sigma}}\|_F &\leq \|\mathbf{\Sigma} - \frac{1}{n}\sum_{i=1}^n \mathbf{N}_{y,i}^\top \mathbf{N}_{y,i}\|_F + \frac{2}{n}\sum_{i=1}^n \|\mathbf{N}_{y,i}^\top (g_{res}(\mathbf{Z}_i) - \hat{g}_{res}(\mathbf{Z}_i))\|_F \\
        &+ \|\frac{1}{n}\sum_{i=1}^n (g_{res}(\mathbf{Z}_i) - \hat{g}_{res}(\mathbf{Z}_i))^\top (g_{res}(\mathbf{Z}_i) - \hat{g}_{res}(\mathbf{Z}_i))\|_F\\
        &\leq A + B + C.
    \end{align}

    We first handle the term $C$:
    \begin{align}
        \mathbb{E}[C] &\leq \frac{1}{n}\sum_{i=1}^n \mathbb{E}\left[\|(g_{res}(\mathbf{Z}_i) - \hat{g}_{res}(\mathbf{Z}_i))^\top (g_{res}(\mathbf{Z}_i) - \hat{g}_{res}(\mathbf{Z}_i))\|_F\right]\\
        &\leq \frac{1}{n}\sum_{i=1}^n \mathbb{E}\left[\|(g_{res}(\mathbf{Z}_i) - \hat{g}_{res}(\mathbf{Z}_i))\|_2^2\right]\\
        &\leq C_1 \kappa_1(n) \hspace{1cm}\text{by assumption $(1)$}.
    \end{align}

    Next, for the term $B$:
    \begin{align}
        \mathbb{E}[B] &\leq \frac{1}{n}\sum_{i=1}^n \mathbb{E}\left[ \|\mathbf{N}_{y,i}^\top (g_{res}(\mathbf{Z}_i) - \hat{g}_{res}(\mathbf{Z}_i))\|_F\right]\\
        &\leq \frac{1}{n}\sum_{i=1}^n \mathbb{E}\left[\|\mathbf{N}_{y,i}\|\right]\mathbb{E}\left[\|(g_{res}(\mathbf{Z}_i) - \hat{g}_{res}(\mathbf{Z}_i))\|_2\right]\\
        &\leq N_{res} \sqrt{C_1} \sqrt{\kappa_1(n)}\hspace{1cm}\text{by assumptions $(1)$ and $(5)$}.
    \end{align}

    For the term $A$, by the Strong Law of Large Numbers, there exists a constant $C_3$ such that:
    \[
    \mathbb{E}[A] \leq \frac{C_3}{\sqrt{n}}.
    \]
    Therefore, we obtain the bound:
    \[
    \mathbb{E}\left[\|\mathbf{\Sigma} - \mathbf{\hat{\Sigma}}\|_{op}\right] \leq N_{res} \sqrt{C_1} \sqrt{\kappa_1(n)} + C_1 \kappa_1(n) + \frac{C_3}{\sqrt{n}}.
    \]
    Similarly, an equivalent reasoning gives:
    \[
    \mathbb{E}\left[\|\mathbf{\Sigma} - \mathbf{\hat{\Sigma}}\|_{op}\right] \leq N_{full} \sqrt{C_2} \sqrt{\kappa_2(n)} + C_2 \kappa_2(n) + \frac{C_4}{\sqrt{n}} \hspace{1cm}\text{using assumptions $(2)$ and $(5)$}.
    \]

    Finally, applying the Davis-Kahan theorem, we have:
    \begin{align}
        \mathbb{E}\left[\|\mathbf{w}_1 - \mathbf{w}_S\|^2_2\right] &\leq \sqrt{2}\frac{\mathbb{E}\left[\|\mathbf{\Sigma}_{\text{res}} - \mathbf{\hat{\Sigma}}_{\text{res}}\|_F\right] + \mathbb{E}\left[\|\mathbf{\Sigma}_{\text{full}} - \mathbf{\hat{\Sigma}}_{\text{full}}\|_F\right]}{\delta_M}\\
        &\leq \sqrt{2}\frac{N_{full} \sqrt{C_2} \sqrt{\kappa_2(n)} + C_2 \kappa_2(n) + \frac{C_4}{\sqrt{n}} + N_{res} \sqrt{C_1} \sqrt{\kappa_1(n)} + C_1 \kappa_1(n) + \frac{C_3}{\sqrt{n}}}{\delta_M}\\
        &= o(\sqrt{\kappa_1(n)} + \sqrt{\kappa_2(n)}),
    \end{align}
    assuming that $\kappa_1(n)$ and $\kappa_2(n)$ decrease no faster than $o(1/n)$, which is typically the case for most of the regression algorithms. This conclude the proof.
\end{proof}

\newpage

\section{Experiments}

\subsection{Simulation experiments}\label{eq:data_generation}
The data are generated according to the following SCM:

\begin{align}\label{eq:simulation_SCM}
\begin{aligned}
    N_x, &N_z \sim \mathcal{N}(0, \mathbf{I}), \\
    N_y &\sim \mathcal{N}(0, \mathbf{\Sigma}), \\
    Z &:= N_z, \\
    X &:= f_a(\mathbf{C}^\top Z)  + N_x, \\
    Y &:= u \mathbf{b}^\top f_a(\Gamma^\top X) + v f_a(\mathbf{D}^\top Z) + w N_y.
\end{aligned}
\end{align}

\paragraph{Causal effect representation}

\begin{figure*}
    \centering
    \includegraphics[width=0.85\linewidth]{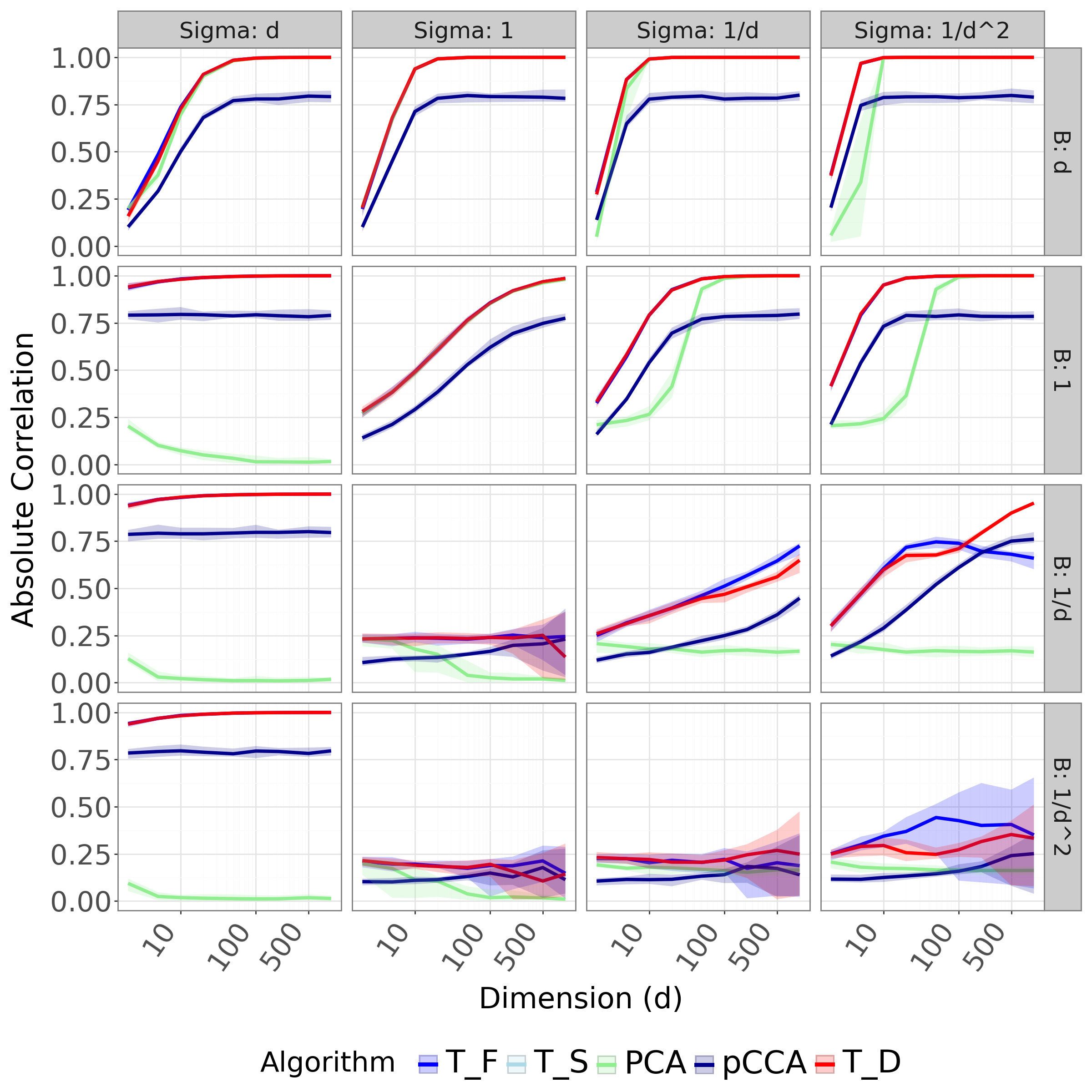}
    \caption{Correlation between $\mathbf{w}^\top Y$ and $\phi(X)$ as $d$ increases. $T_D$ consistently outperforms all methods, recovering $\phi(X)$ as $d$ grows, provided that $\mathbf{b}$ faster than $\mathbf{\Sigma}$. Columns are indexed by as A, B, C, D and rows by $1, 2, 3, 4$. }
    \label{fig:DR_noise_behavior_Noise}
\end{figure*}

\begin{figure*}
    \centering
    \includegraphics[width=0.75\linewidth]{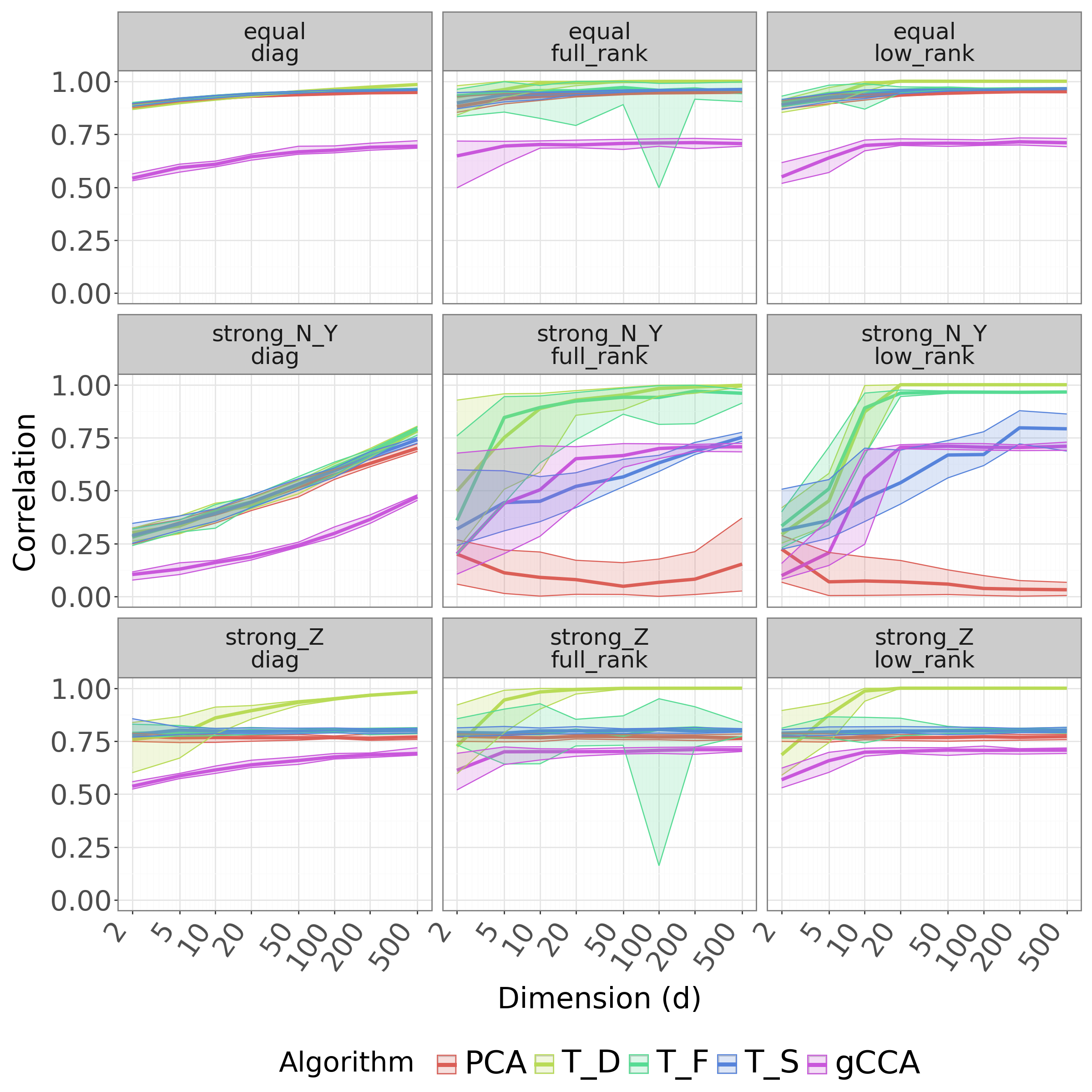}
    \caption{Experiments with different noise structure ($\mathbf{\Sigma}$ being diagonal, full rank and low rank) and scaling factors ($(u, v, w)$ as $(1/3, 1/3, 1/3)$, $(0.1, 0.1, 0.8)$, and $(0.1, 0.8, 0.1)$ for equal, Strong\_N\_Y and Strong\_Z). Overall, learning algorithm $T_D$ performs better and tends to converge.}
    \label{fig:DR}
\end{figure*}

Comparing how the different learning algorithms behave in different noise contexts seems relevant. Primarily, we can observe in the setting Strong\_N\_Y low\_rank (Fig. \ref{fig:DR}) that the increase in performance using $T_F$ and $T_D$ is due to the low-rank structure of the noise. The overall better performance of $T_D$ over $T_F$ and pCCA is due to the correlation between $X$ and $Z$.

To better understand the behaviour of our algorithms, we conducted experiments in two additional settings:

\begin{enumerate}
    \item \textbf{High-dimensional setting:} We conducted a similar experiment under different conditions on $\mathbf{b}$ and $\mathbf{\Sigma}$, increasing the dimensionality. In this case, however, we significantly reduced the sample size to $n=100$, such that as $d$ grows, we obtain $n<d$.
    \item \textbf{Nonlinear setting:} Again, we conducted a similar experiment with different conditions on $\mathbf{b}$ and $\mathbf{\Sigma}$. Still, here we applied a nonlinear mapping $f_a(z):= \exp(-z^2/2)\sin(az)$ with $a \in \{1, 2, 3\}$. We use a random forest algorithm with 100 trees as an estimator of the conditional expectation.
    \item \textbf{$X$ independent of $Z$ setting:} We conduct experiments to clarify the discrepancy of performance between pCCA and $T_D$ by generating the data such that $X$ and $Y$ are independent.
\end{enumerate}

\begin{figure*}
    \centering
    \includegraphics[width=0.85\linewidth]{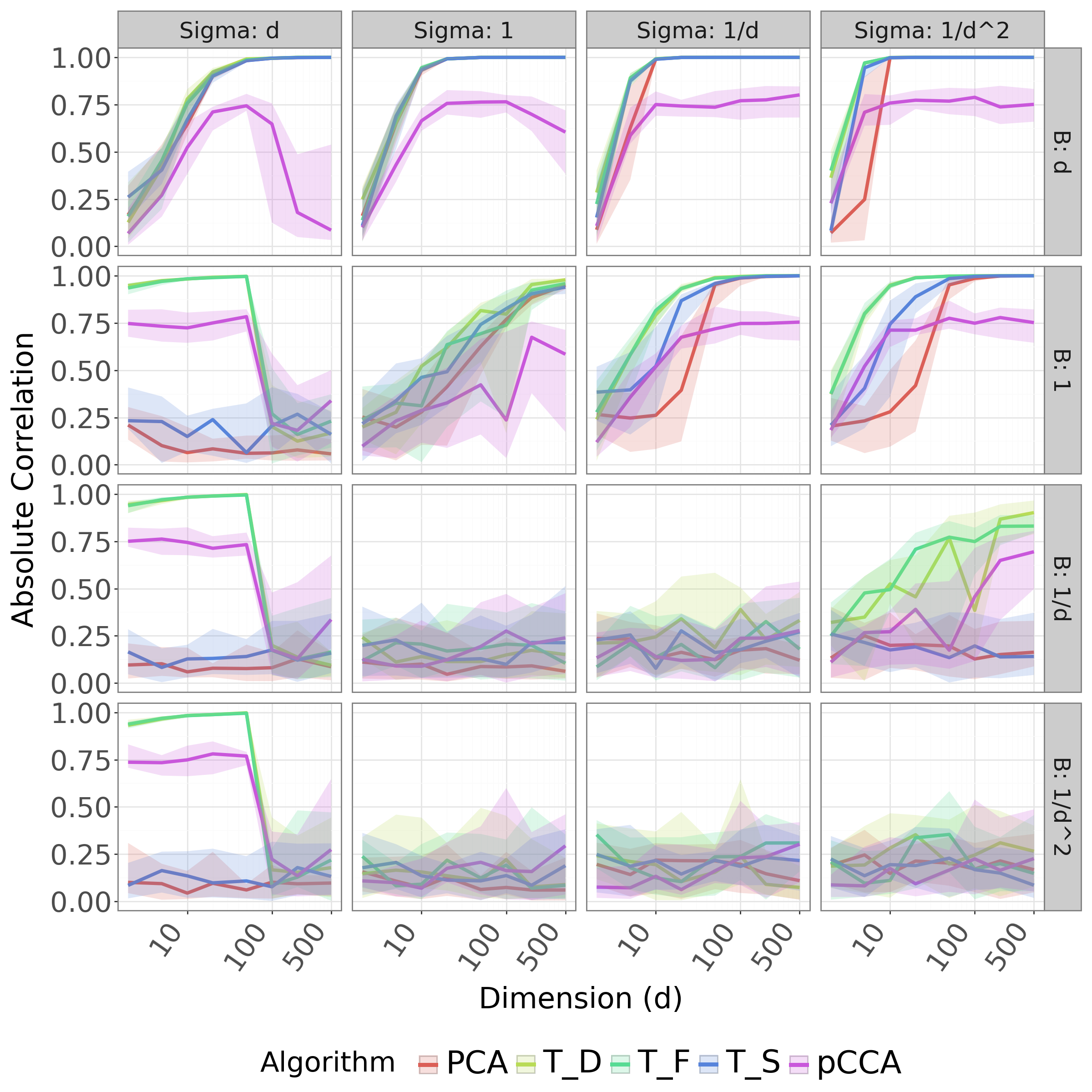}
    \caption{High dimensional experiment using $100$ samples for training. As observed in the last row, all algorithms fail to recover the signal when the noise variance increases rapidly and the number of samples falls below the outcome dimension.
}
    \label{fig:DR_high_dimensional}
\end{figure*}

In the high-dimensional setting, as shown in Fig. \ref{fig:DR_high_dimensional}, we observe results that are very similar to those in the large-sample setting, with one key difference: when $\mathbf{\Sigma}$ increases rapidly with $d$ (row 1), and $d > n$, the model performance drops significantly to near zero and when $\mathbf{b}$ is growing to slowly compared to $\sigma$ ($\sigma=[1, 1/d]$ and $\mathbf{b}=[1/d, 1/d^2]$). It would be interesting to further evaluate how the regularisation parameter (see Section \ref{sec:stability}) might improve performance in this specific case. However, in this setting, ensuring algorithmic convergence is particularly challenging, as the signal strength is constrained by the number of available samples. As a result, the SNR is unlikely to grow unbounded unless the inherent noise in the data is minimal.

Interestingly, in the nonlinear setting, the learning algorithm $T_D$ is still able to recover $\phi(X)$ in most cases. In contrast, other algorithms show greater difficulty in achieving convergence under these conditions. This highlights the potential of learning algorithm $T_D$ to recover direct effects, even in complex nonlinear settings effectively.

\begin{figure*}
    \centering
    \includegraphics[width=0.85\linewidth]{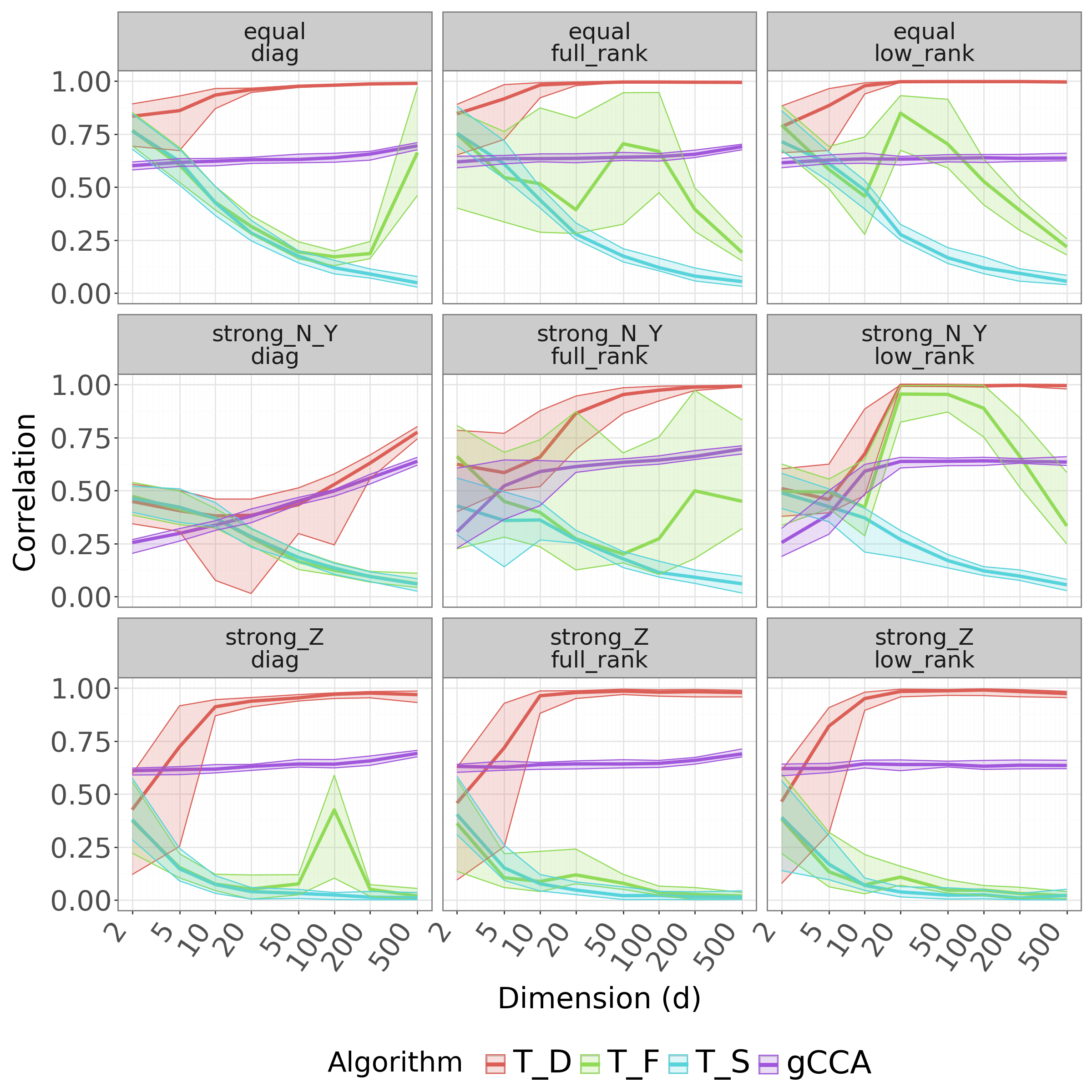}
    \caption{Experiments with nonlinear map $f_a$ and with different noise structure ($\mathbf{\Sigma}$ being diagonal, full rank and low rank) and scaling factors ($(u, v, w)$ as $(1/3, 1/3, 1/3)$, $(0.1, 0.1, 0.8)$, and $(0.1, 0.8, 0.1)$ for equal, Strong\_N\_Y and Strong\_Z). Overall, the learning algorithm $T_D$ performs better and tends to converge.}
    \label{fig:DR_nonlinear}
\end{figure*}

We also conducted an experiment where $X$ and $Z$ were generated as independent variables to highlight the potential advantages of our learning algorithms. In this case, we observed that partial CCA (pCCA) could recover the latent structure similarly to the other algorithms. This result highlights the robustness of $T_D$, as it remains stable across different structural relationships between $X$ and $Y$, reinforcing its applicability in both confounded and mediated settings.

\begin{figure*}
    \centering
    \includegraphics[width=0.85\linewidth]{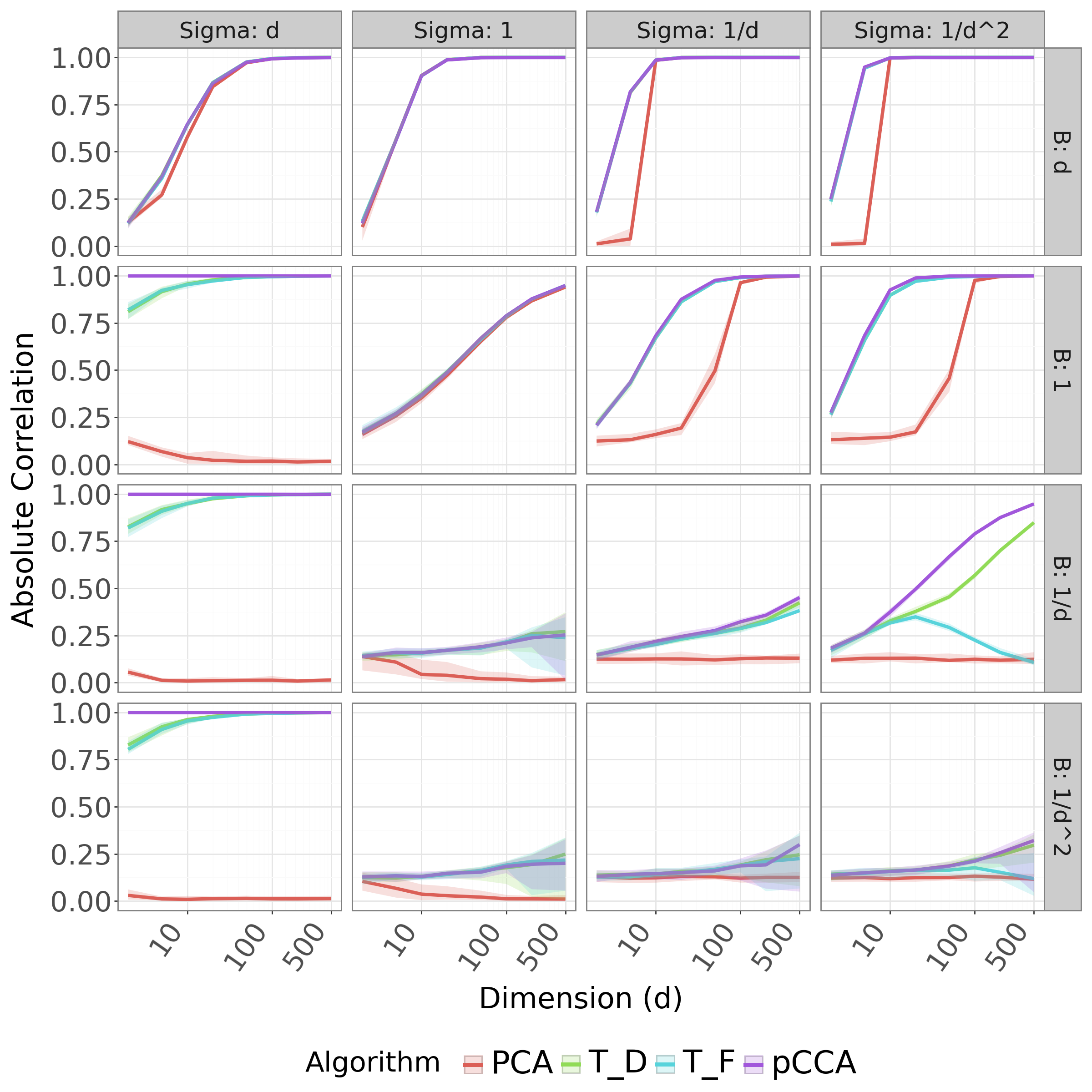}
    \caption{Experiment where $X \indep Z$. When the dependence between $X$ and $Z$ is removed, the pCCA algorithm performs similarly to $T_D$. This highlights the effectiveness of our approach in scenarios with confounding or mediation effects.}

    \label{fig:DR_noise_behavior_indep}
\end{figure*}

\paragraph{Hypothesis testing}

\begin{figure}
    \centering
    \includegraphics[width=0.9\linewidth]{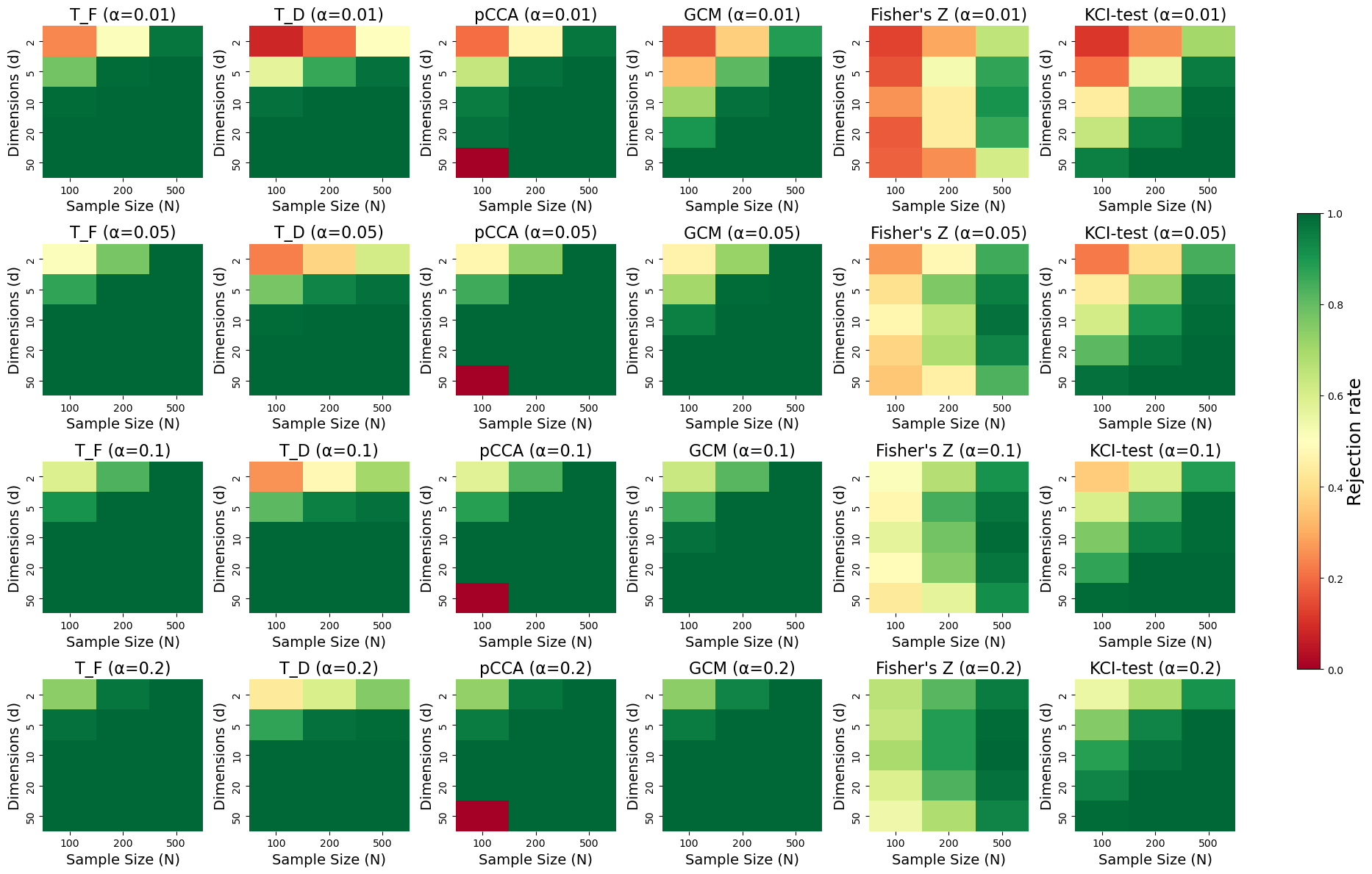}
    \caption{Power of the different methods. $T_D$, $T_F$, and pCCA generally exhibit better performance compared to the other approaches. This is partly because they rely on linear Gaussian models, which constrain the alternative hypotheses, thereby improving the power of these tests.}

    \label{fig:power_all}
\end{figure}

\begin{figure}
    \centering
    \includegraphics[width=0.9\linewidth]{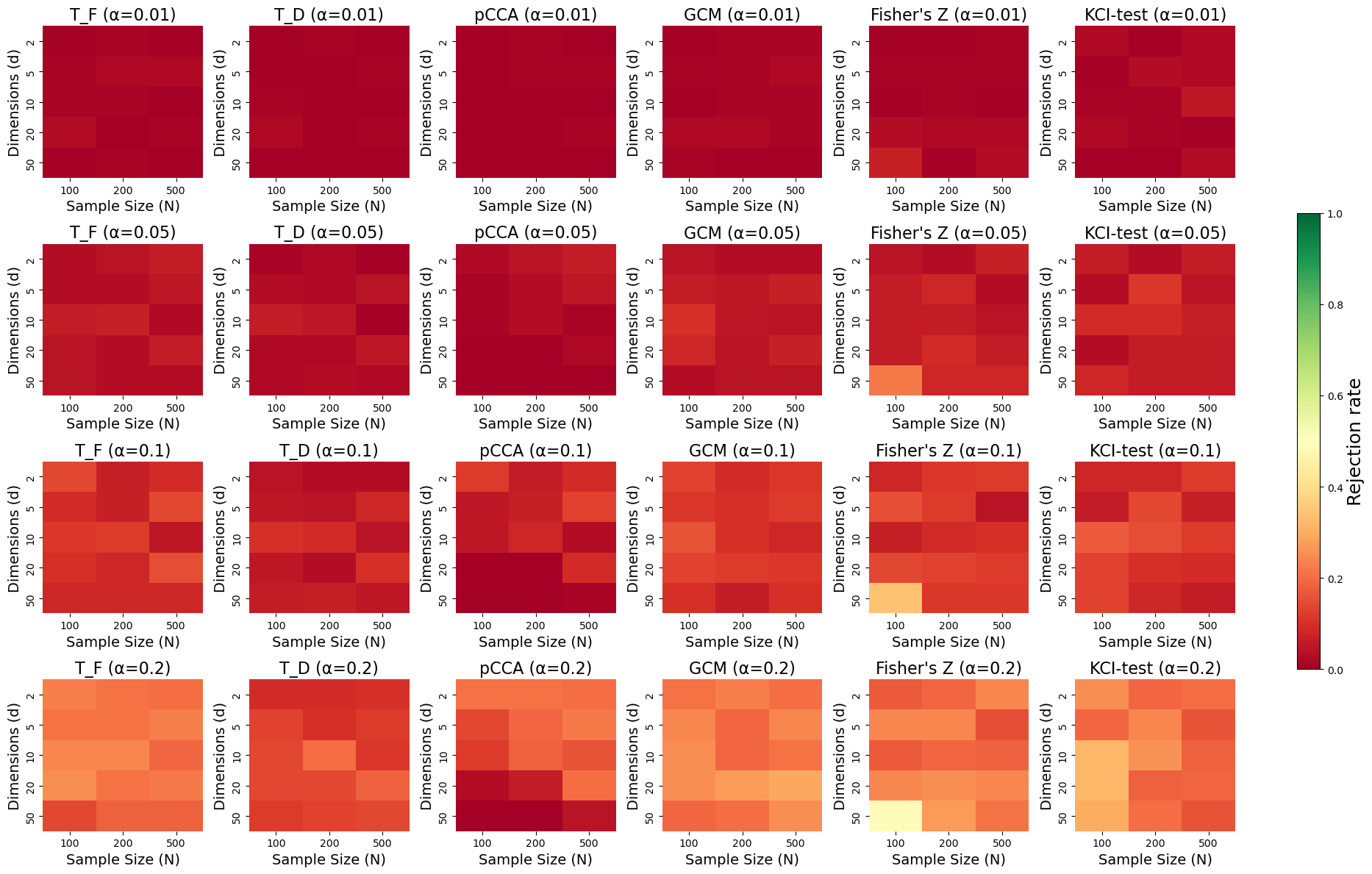}
    \caption{Type $I$ error control. We can observe that all methods have a good control of type $I$ error except the Fisher $Z$ test, which has a poor control in low sample high dimensional settings.}
    \label{fig:type_I_control}
\end{figure}

For a test to be valid, it must control the Type I error rate. Specifically, if we test at level $\alpha$, then under the null hypothesis $H_0$, the probability of rejecting $H_0$ should be less than or equal to $\alpha$. In Fig. \ref{fig:type_I_control}, we observe that all the methods control the Type I error reasonably accurately.

\vfill

\newpage

\subsection{Real-World experiments}

\paragraph{Separating internal climate variability from the externally forced response.}

\begin{figure}\label{fig:mse_boxplot}
    \centering
    \includegraphics[width=1\linewidth]{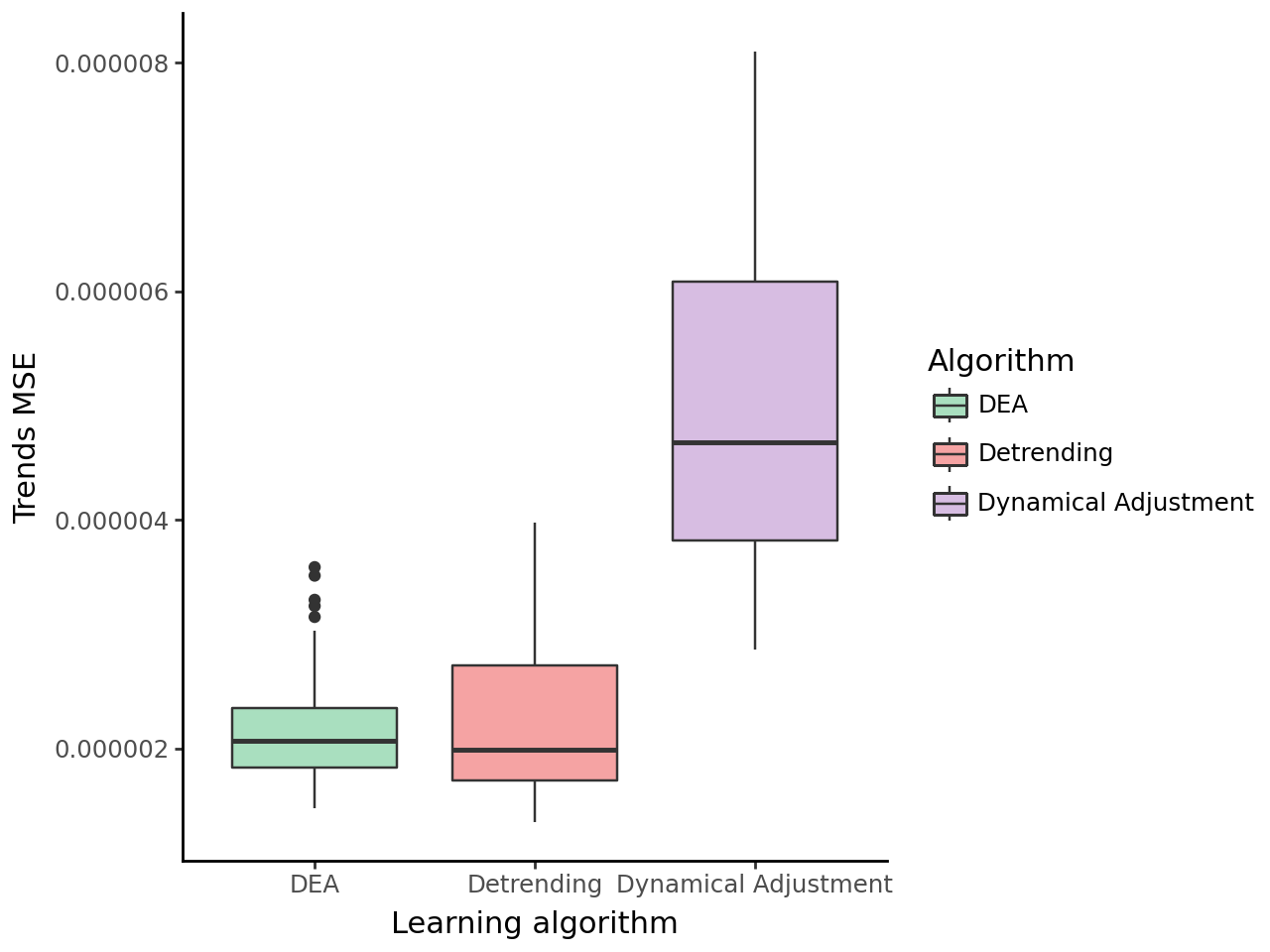}
        \caption{(a) Forced response trends. Mean Squared Error (MSE) of different algorithms in reconstructing (a) internal climate variability trends and (b) forced response trends. The Direct Effect Analysis (DEA) algorithm, using the basis $(\mathbf{\Sigma}^{-1}\mathbf{b}, \mathbf{b}^\perp)$, i.e. $T_D$ algorithm, is compared to Detrending and Dynamical Adjustment (two most common approaches for separating internal from external climate variability). Overall, DEA and Detrending perform better. DEA outperforms Detrending for internal variability trend estimation but has a higher median MSE for forced trend reconstruction. However, DEA provides better worst-case control in this case.}
\end{figure}

\begin{figure}
        \centering
        \includegraphics[width=1\linewidth]{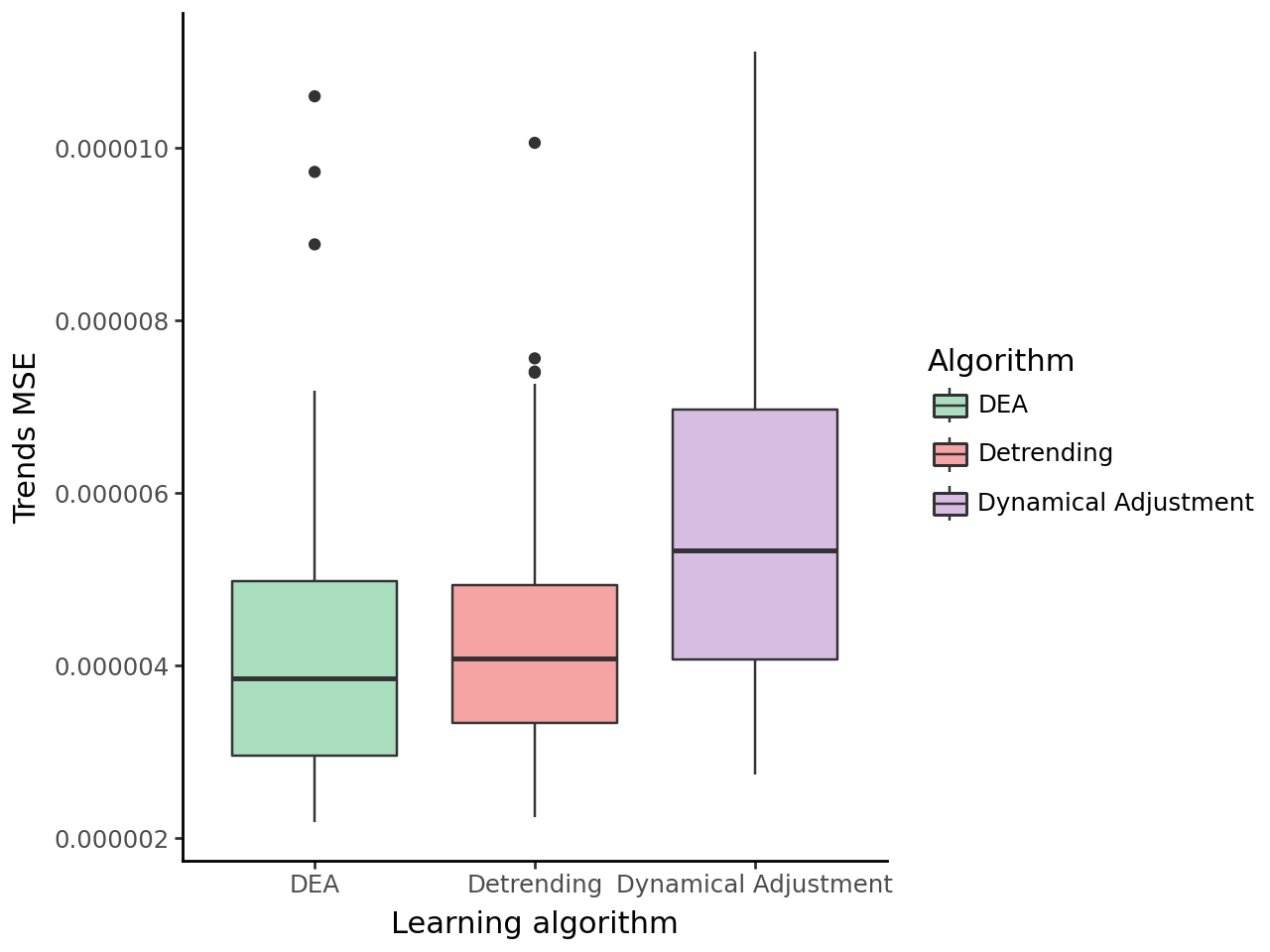}
        \caption{(b) Internal variability trends}
\end{figure}


This experiment aims to assess the performance of our learning algorithms in disentangling internal climate variability from the forced response to external factors, such as greenhouse gas (GHG) emissions or solar radiation. For this analysis, we focus on temperature fields. We use $M = 50$ members from the CESM2 historical climate simulations \citep{danabasoglu2020community}, covering 1880 to 2014. The variables under consideration are Sea Level Pressure (SLP) and Temperature (T), with monthly data yielding $1669$ samples per member. The detrended SLP field is treated as a proxy for internal variability ($Z \in \mathbb{R}^{648}$). In contrast, the temperature field ($Y \in \mathbb{R}^{648}$) serves as the response variable of interest. The temperature for member $i$, at location $j$, and time $t$, is denoted by $Y^{(i)}_j(t)$.

As a proxy for climate external forcing, we use a smoothed version (5-year moving average) of the Global Mean Temperature (GMT), which is computed as a spatial average of the temperature field ($X \in \mathbb{R}$):
\begin{equation}
    X(t) = \frac{1}{\text{years} \times 12} \sum_{\tau=1}^{\text{years} \times 12} \frac{1}{d} \sum_{j=1}^d Y_j(t - \tau)
\end{equation}
where $d$ represents the number of spatial locations.

The climate-forced response, $Y_{\text{forced}}$, is calculated as the ensemble mean over all simulation members, $Y^{(i)}$:
\begin{equation}\label{eq:intern_forced}
    Y_{\text{forced}, j}(t) = \frac{1}{M} \sum_{i=1}^M Y^{(i)}_j(t) \quad \text{and} \quad Y_{\text{internal}, j}^{(i)}(t) = Y^{(i)}_j(t) - Y_{\text{forced}, j}(t),
\end{equation}
where $Y_{\text{internal}}$ represents the true internal variability of $Y$ after removing the climate-forced component.

The Direct Effect (DEA) algorithm (employing $T_D$) is applied as follows: We train DEA using the triplet $\{GMT(t), T(t), SLP(t)\}_{t=1}^{years\times 12}$ as realisations of $(X, Y, Z)$, where $GMT$ serves as the climate external forcing proxy ($X$), $T$ represents the temperature response variable, and $SLP$ is a proxy for the internal climate variability ($Z$). 
Once the model is trained, we project the data onto the null space of the vector $\mathbf{b}$, denoted as $\mathbf{b}^\perp$, to recover the internal variability component $\hat{Y}_{\text{internal}}$. This projection isolates the portion of the temperature field that is not correlated with the external forcing, allowing us to separate the forced and internal components effectively. Finally, we compute the climate-forced response as $\hat{Y}_{\text{forced}} = Y - \hat{Y}_{\text{internal}}$, which provides an estimate of the temperature response attributed to external forcing alone.

We compare our learning algorithm with two common approaches used in climate science:

\begin{enumerate}
    \item \textbf{Detrending:} A simple linear model predicts $Y$ from $X$, providing an estimate of the climate-forced response, $\hat{Y}_{\text{forced}}$, and the dynamical component as $\hat{Y}_{\text{internal}} = Y - \hat{Y}_{\text{forced}}$.
    \item \textbf{Dynamical Adjustment} \citep{Sippel2019}: A model is trained using both $X$ (GMT) and $Z$ (SLP). Predictions are made by setting $Z$ to zero, isolating the dynamical component.
\end{enumerate}

\begin{figure}
    \centering
    \includegraphics[width=1\linewidth]{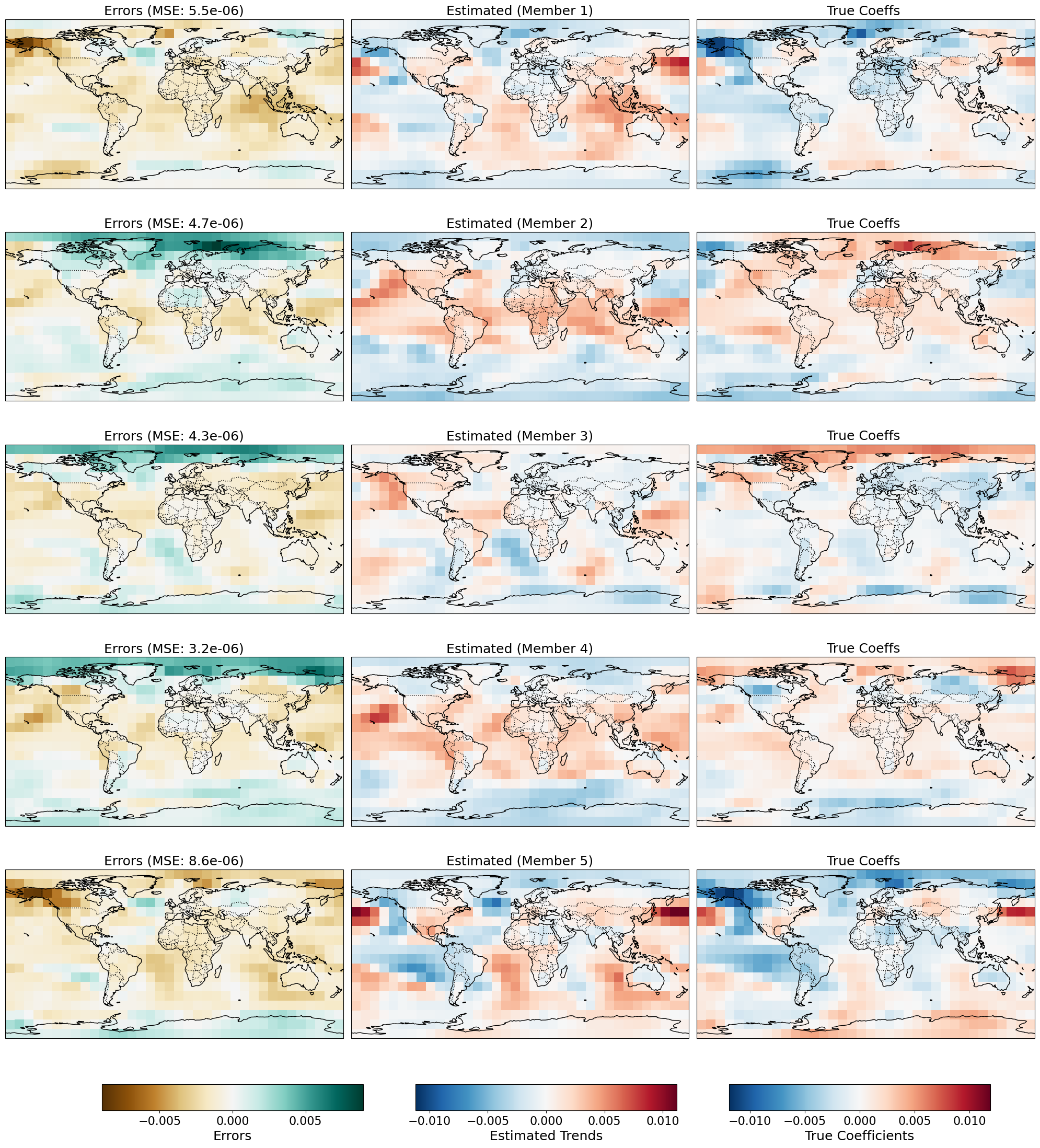}
    \caption{Trends of the reconstructed internal climate variability using the DEA algorithm. The algorithm captures general warming and cooling patterns but underestimates trends in the North Pole and overestimates them in Western America.}
    \label{fig:trends_maps_DEA}
\end{figure}

\begin{figure}
    \centering
    \includegraphics[width=1\linewidth]{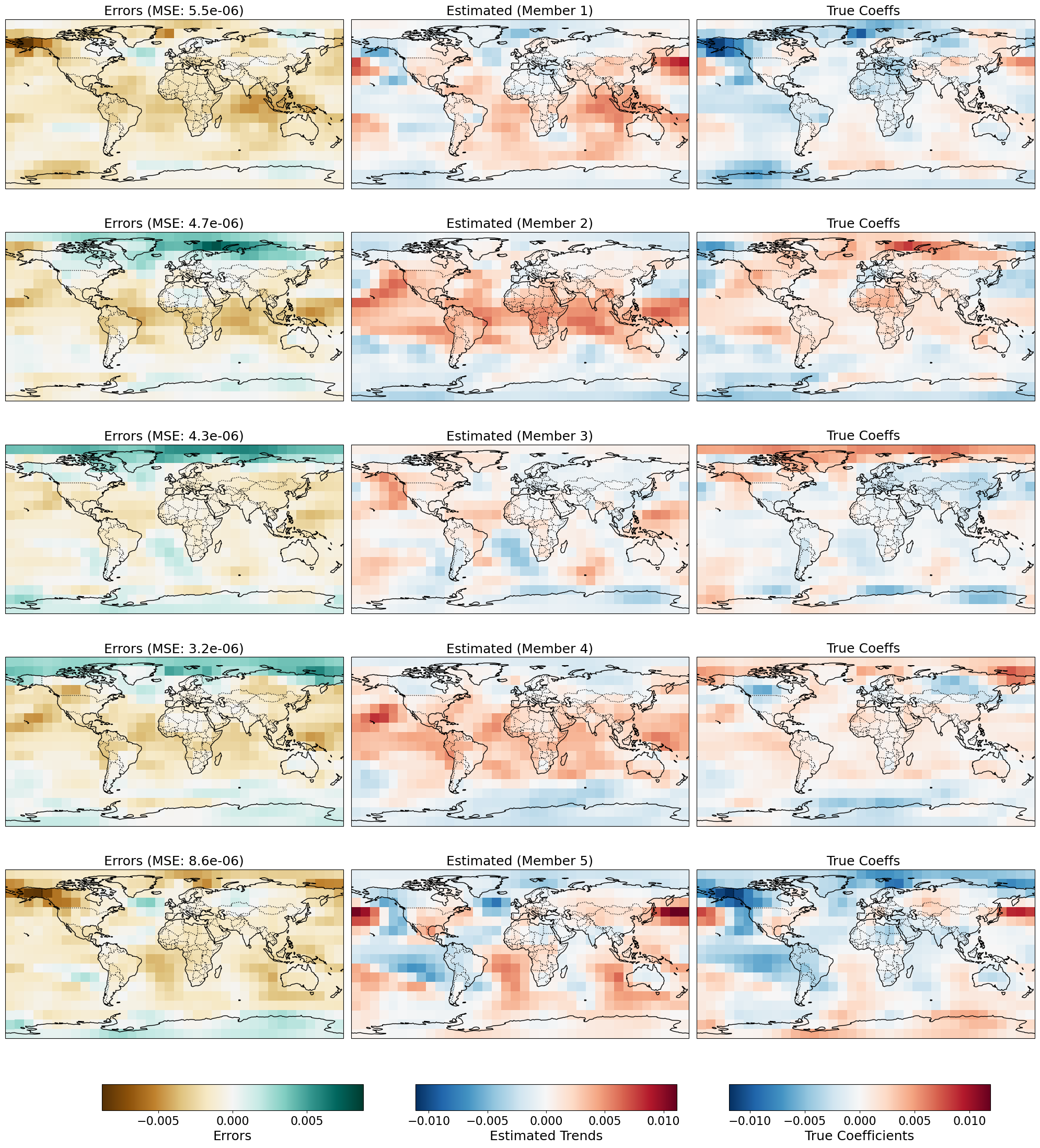}
    \caption{Trends of the reconstructed internal climate variability using the Detrending algorithm. The algorithm captures general warming and cooling patterns but underestimates trends in the poles and overestimates trends in Western America and Indonesia.}
    \label{fig:trends_maps_Detrending}
\end{figure}

\begin{figure}
    \centering
    \includegraphics[width=0.95\linewidth]{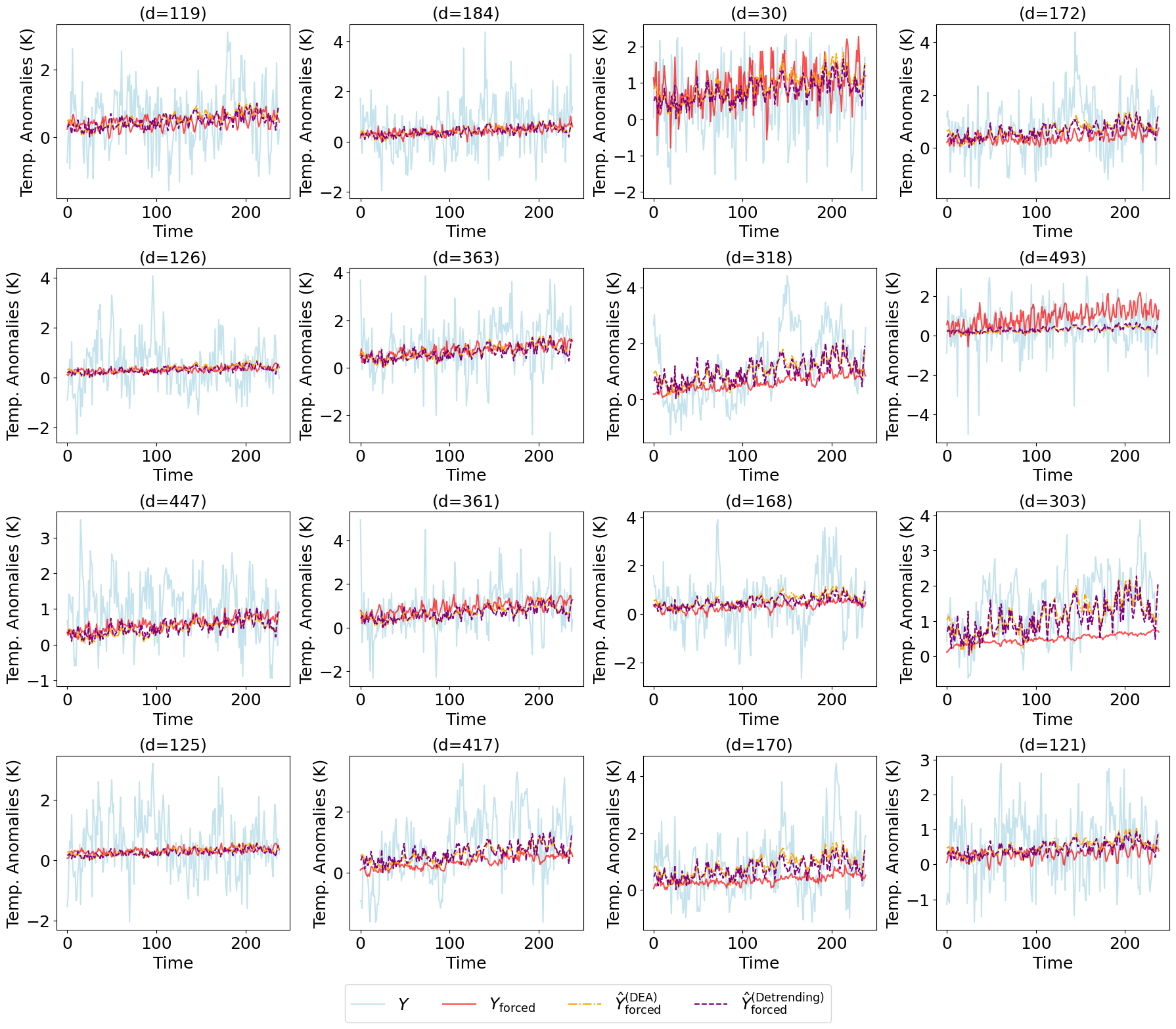}
    \caption{Comparison of original observations $Y$ and the reconstructed climate-forced response $\hat{Y}_{\text{forced}}$ at 16 randomly selected locations for both DEA and Detrending.}
    \label{fig:climate_experiment_forced_response_trends_TS}
\end{figure}

We compute trends for the dynamical components ($Y_{\text{internal}}$, $\hat{Y}_{\text{internal}}$, $Y_{\text{forced}}$, and $\hat{Y}_{\text{forced}}$) over 20-year periods. The performance of the methods is evaluated using the following metrics:

\begin{itemize}
    \item \textbf{20-year trends MSEs:} MSEs for the trends of the three methods are shown in Fig. \ref{fig:mse_boxplot} (a) for forced trends and (b) for internal variability trends. The boxplots display the MSE distributions across different simulation members.
    \item \textbf{20-year trends maps (internal variability):} Internal variability trends are compared spatially in Fig. \ref{fig:trends_maps_DEA} for DEA and Fig. \ref{fig:trends_maps_Detrending} for Detrending to better understand model biases.
    \item \textbf{20-year trends time series (forced response):} Forced response trends are compared over time, with time series for DEA and Detrending plotted for randomly selected locations in Fig. \ref{fig:climate_experiment_forced_response_trends_TS}.
\end{itemize}

We train the algorithms (DEA, Detrending, Dynamical Adjustment), extract latent structures $\hat{Y}_{\text{forced}} = \mathbf{w}^\top Y$, and compare the trends of the last 20 years of $\hat{Y}_{\text{forced}}$ and $Y_{\text{forced}}$. Figure \ref{fig:mse_boxplot} shows that DEA performs similarly to Detrending but outperforms Dynamical Adjustment.

A qualitative evaluation of the trend maps generated by DEA (Figure \ref{fig:trends_maps_DEA}) shows that the algorithm captures the warming and cooling patterns. However, both DEA and Detrending tend to underestimate trends in polar regions where temperature trends are generally stronger.

Figure~\ref{fig:climate_experiment_forced_response_trends_TS} shows that both Detrending and DEA effectively capture the forced response trends. However, in regions where the forced response exhibits high variability (e.g., at $d=30$ or $d=493$, typically located in polar regions), both methods struggle to fully capture this variability.
 This may be due to the smoothing of GMT in the external forcing, but this observation warrants further investigation, as these regions of high variability may also reflect model artefacts. Further exploration of these phenomena is needed.

\end{document}